\documentclass{article}

 \usepackage[main, final]{neurips_2026}

\usepackage{microtype}
\usepackage{graphicx}
\usepackage{subcaption}
\usepackage{booktabs} 
\usepackage{svg} 
\usepackage{graphicx}
\usepackage{amsmath}
\usepackage{mathtools}
\usepackage{xcolor}
\usepackage{enumitem}
\usepackage{booktabs}  
\usepackage{multirow}  
\usepackage{amsmath}   
\usepackage{makecell}  
\usepackage{subcaption}   
\usepackage{float}
\usepackage[table]{xcolor} 
\usepackage{natbib}
\usepackage{authblk}

\definecolor{lightlayer}{rgb}{240, 248, 255}
\definecolor{aliceblue}{RGB}{240, 248, 255}
\definecolor{mintgreen}{RGB}{245, 255, 250}
\definecolor{lpink}{RGB}{255, 240, 245}

\newcommand{\best}[1]{\textbf{#1}}
\newcommand{\second}[1]{\underline{#1}}

\usepackage{tcolorbox}
\tcbuselibrary{skins} 
\tcbset{
  takeawaybox/.style={
    enhanced,
    colframe=gray!60!black,  
    colback=gray!5!white,   
    boxrule=0.4pt,          
    arc=1pt,                
    left=8pt, right=8pt, top=5pt, bottom=5pt, 
    fontupper=\normalsize\itshape, 
    after skip=5pt, before skip=5pt, 
  }
}

\usepackage{hyperref}
\usepackage[capitalize]{cleveref}

\usepackage[utf8]{inputenc} 
\usepackage[T1]{fontenc}    
\usepackage{hyperref}       
\usepackage{url}            
\usepackage{booktabs}       
\usepackage{amsfonts}       
\usepackage{nicefrac}       
\usepackage{microtype}      
\usepackage{xcolor}         

\title{Unveiling Implicit Advantage Symmetry: Why GRPO Struggles with Exploration and Difficulty Adaptation}

\author[1]{Zhiqi Yu\textsuperscript{*}}
\author[2]{Zhangquan Chen\textsuperscript{*}}
\author[3]{Mengting Liu}
\author[3]{Heye Zhang}
\author[1]{Liangqiong Qu\textsuperscript{$\dagger$}}

\affil[1]{School of Computing and Data Science, The University of Hong Kong}
\affil[2]{Tsinghua Shenzhen International Graduate School, Tsinghua University}
\affil[3]{School of Biomedical Engineering, Sun Yat-sen University}

\affil[ ]{\small
\textsuperscript{*}Equal contribution. \quad
\textsuperscript{$\dagger$} Corresponding author: \texttt{liangqqu@hku.hk}
}

\affil[ ]{\small
\texttt{zhiqiyu777@connect.hku.hk} \quad
\texttt{czq23@mails.tsinghua.edu.cn} \quad
\texttt{liangqqu@hku.hk}
}

\affil[ ]{\small
\texttt{\{liumt55, zhangheye\}@mail.sysu.edu.cn}
}

\begin{document}

\maketitle

\begin{abstract}
   Reinforcement Learning with Verifiable Rewards (RLVR), particularly GRPO, has become the standard for eliciting LLM reasoning. However, its efficiency in exploration and difficulty adaptation remains an open challenge. In this work, we identify an implicit advantage symmetry inherent in Group Relative Advantage Estimation (GRAE) as a structural property that provides a new perspective for understanding these bottlenecks. This symmetry induces two critical limitations: (i) at the group level, strict symmetry in weights between correct and incorrect trajectories leaves unsampled action logits unchanged, thereby hindering the exploration of novel correct solution. (ii) at the sample level, the algorithm implicitly prioritizes medium-difficulty samples, remaining agnostic to the non-stationary demands of difficulty focus. Through controlled experiments, we reveal that this symmetric property is sub-optimal, yielding two pivotal insights: (i) asymmetrically down-weighting the advantages of correct trajectories encourages essential exploration but risks instability; (ii) learning efficiency can be boosted by a curriculum-like transition—prioritizing simpler samples initially before gradually shifting to complex ones. Motivated by these findings, we propose Asymmetric GRAE (A-GRAE), which dynamically modulates exploration incentives and sample-difficulty focus. Experiments across seven benchmarks demonstrate that A-GRAE consistently improves GRPO and its variants across both LLMs and MLLMs. Code is available at \url{https://github.com/HKU-HealthAI/A-GRAE}.
\end{abstract}

\section{Introduction}

Reinforcement Learning with Verifiable Rewards~(RLVR) \citep{ouyang2022training,achiam2023gpt,shao2024deepseekmath,wen2025reinforcement} has become a cornerstone for activating Foundation Models’ capacity to address complex reasoning tasks through Chain-of-Thought (CoT) generation. Among various algorithms, Group Relative Policy Optimization (GRPO) has emerged as the standard implementation in advanced systems such as DeepSeek‑R1~\citep{guo2025deepseek} and OpenAI‑o1~\citep{jaech2024openai}. The core innovation of GRPO lies in Group Relative Advantage Estimation (GRAE), which computes relative advantages within sampled groups to obviate the need for a value model.

\begin{figure*}[t]
\centering
\includegraphics[width=0.93\linewidth]{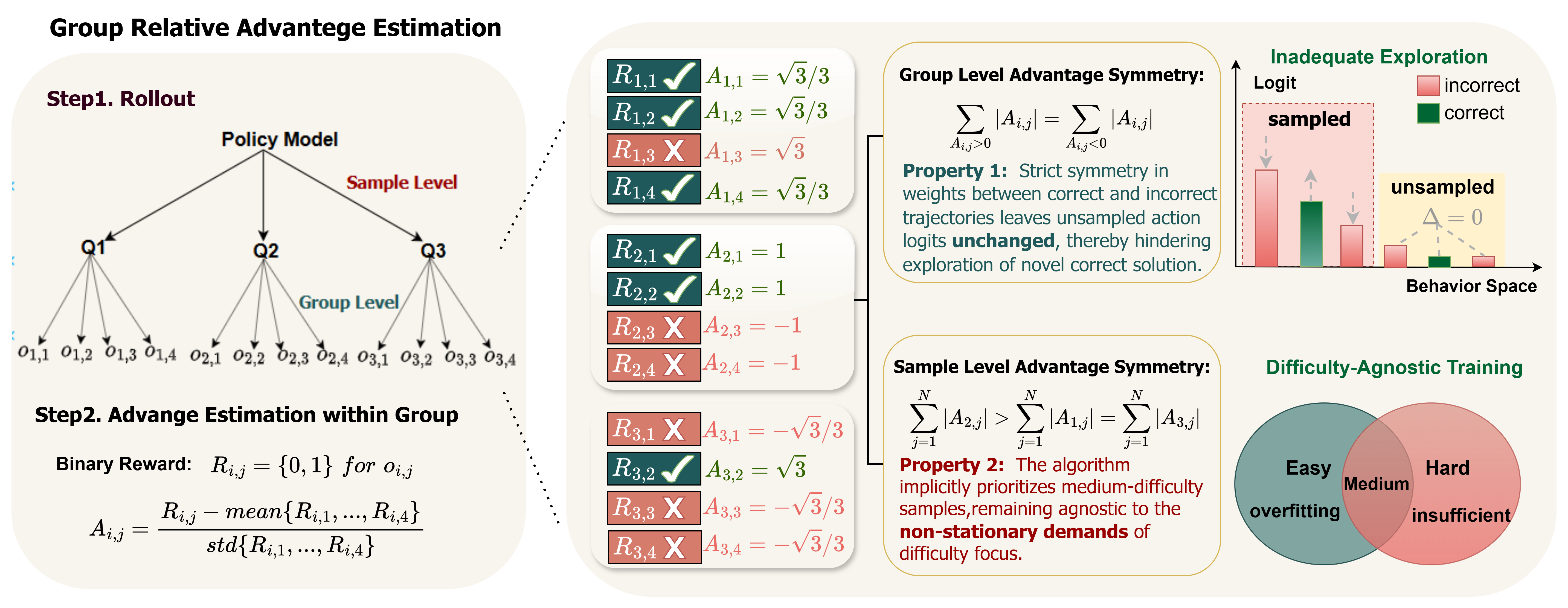}
\caption{\textbf{The two-fold implicit advantage symmetry problem of GRAE in GRPO.} At the group level, the advantage weights for correct trajectories equal those of incorrect trajectories. This symmetry leads to the logits of low-probability correct paths unchanged under the advantage-driven update, thereby hindering the model's exploration. At the sample level, medium-difficulty samples exhibit the largest sum of absolute advantage values, leading to insufficient training on harder data.}
\label{fig:framework}
\vspace{-5pt}
\end{figure*}

Recent studies have identified two critical limitations of GRPO.
The first is \textbf{capability boundary shrinkage}:
 A prominent critique posits that GRPO primarily leads to an exploration-exploitation trade-off within existing policy constraints rather than effectively expanding the decision boundary~\citep{yue2025does, bamba2025xrpo, wang2026targeted, ma2025learning,huang2025beyond}. This claim is supported by empirical evidence showing that GRPO’s Pass@$k$ can even fall below that of the base model at large $k$.  This phenomenon has motivated approaches to improves Pass@$k$ based on negative reinforcement, entropy-aware optimization, and Pass@$k$-oriented objectives
\citep{zhu2025surprising,yao2025debate,chen2025pass}.
The second limitation is \textbf{inadequate focus on problem difficulty}: GRPO’s reward mechanism is difficulty-agnostic, treating all tasks uniformly without accounting for their inherent complexity or the model’s current capacity. To this end, recent methods\citep{zhang2025grpo,zhou2025codapo,zhang2025clpo,dai2026harder} have sought to improve training efficiency and accuracy by either reweighting hard or informative samples or incorporating curriculum-style strategies.

However, these two lines of work remain largely separate, and few approaches achieve substantial gains in both accuracy and Pass@$k$ due to the fundamental conflict between sampling efficiency and diversity. This naturally raises a key question: \textit{Can a unified perspective explain these two limitations of GRPO and, in turn, guide a principled way to overcome them at the same time?}


In this work, we show that the two deficiencies can be understood through a previously overlooked \textbf{implicit advantage symmetry inherent in GRAE}.
As illustrated in~\cref{fig:framework}, this symmetry appears at two complementary levels. At the group level, the advantage weights for correct and incorrect trajectories are strictly equivalent. Through a behavior-space logit analysis, we show that this symmetry implies that an unsampled trajectory receives no direct reward-specific advantage signal, thereby limiting reward-directed exploration of potentially correct but unseen solutions. At the sample level, by quantifying the absolute sum of advantages across samples, we reveal that the algorithm implicitly prioritizes medium-difficulty instances. Consequently, the optimization remains agnostic to the non-stationary demands of training dynamics, failing to adapt its focus as the model evolves.

On this theoretical ground, we perform controlled interventions that deliberately break GRAE's intrinsic advantage symmetry at both the group and sample levels to examine its causal effect on learning dynamics. 
Our experiments reveal that the symmetric property is sub-optimal and yield two key design principles: (i) asymmetrically down-weighting the weights of correct trajectories fosters essential exploration; (ii) learning efficiency can be boosted by a curriculum-like progression, 
which prioritizes simpler samples initially before escalating to complex ones. Motivated by these findings, we propose \textbf{Asymmetric GRAE (A-GRAE)}, which refines the original GRAE strategy by explicitly instantiating these two principles within the GRPO framework: it introduces asymmetric exploration to push the policy beyond its current solution set, and a curriculum-like progression learning schedule to align the optimization focus with the model’s evolving capability.

To fully validate A-GRAE, we conduct comprehensive experiments across seven diverse benchmarks—spanning both natural language reasoning and vision-language reasoning tasks—using various commonly used LLMs and VLMs. Experimental results prove that A-GRAE consistently enhances the performance of GRPO and its representative variants DAPO~\cite{yu2025dapo} and Dr.GRPO~\cite{liu2025understanding} across all settings, with significant improvements in key metrics (e.g., accuracy, Pass@$k$).

In summary, our contributions are as follows:
\begin{itemize}
    \item We identify \textbf{advantage symmetry} as a two-level structural property of GRAE, providing a unified perspective that connects previously separate methods for exploration and difficulty-aware learning in GRPO-based RLVR.

    \item Based on this unified perspective, we systematically break group-level and sample-level symmetry through controlled experiments and uncover two distinct findings beyond existing fixed strategies: excessive group-level exploration can cause training instability, while the preferred sample difficulty dynamically shifts from easier to harder queries during training.

    \item Guided by these findings, we propose \textbf{A-GRAE}, which jointly and dynamically adjusts group-level exploration and sample-level difficulty focus. Extensive experiments across seven benchmarks, multiple model families, and several GRPO variants demonstrate consistent improvements in both accuracy and Pass@$k$.
\end{itemize}

\section{Preliminary}


\textbf{Notation.} In this work, we frame a foundation model (LLM or MLLM), parameterized by $\theta$, as a policy model: $\pi_\theta$ and $\pi_{\theta_{\text{old}}}$ denote the current and prior policies, respectively. For batch processing, we sample a  question $q$ from the dataset $\mathcal{Q}$. For each question $q$, the policy generates a set of $G$ candidate responses $\{o_{i}\}_{i=1}^{G}$. Each question-response pair $(q, o_{i})$ is then assigned a scalar reward $r_{i}$ via a rule-based verifier, where $i \in \{1,2,\dots,G\}$ indexes responses per question. By default, we employ a binary accuracy reward: $r_{i} = 1$ if response $o_{i}$ is correct, and $r_{i} = 0$ otherwise. For each response $o_{i}$, we further compute a group-relative advantage value $A_{i}$ for policy optimization.

\textbf{Group Relative Policy Optimization (GRPO).} GRPO simplifies the training pipeline of PPO by removing the critic model—typically matching the size of the policy model—and instead estimates baselines via group-level reward scores. Specifically, for each query $q$, GRPO samples a set of responses $\{o_1, o_2, \dots, o_G\}$ from the prior policy $\pi_{\theta_{\text{old}}}$ and optimizes the target policy $\pi_\theta$ by maximizing the following objective function:
\begin{equation}
\label{eq:grpo}
\begin{aligned}
J_{\text{GRPO}}(\pi_\theta) = \mathbb{E}_{q \sim \mathcal{Q}, \{o_i\}_{i=1}^G \sim \pi_{\theta_{\text{old}}}(\cdot|q)}\frac{1}{G} \sum_{i=1}^G \frac{1}{|o_i|}
\sum_{t=1}^{|o_i|}  \Bigl[\min\bigl( \rho_{i,t} A_{i,t}, \operatorname{clip}\left( \rho_{i,t}, 1\pm\epsilon \right) A_{i,t} \bigr) - \beta \mathbb{D}_{\text{KL}} \Bigr],
\end{aligned}
\end{equation}
where $\rho_{i,t}=\frac{\pi_\theta(o_{i,t}|q,o_{i,<t} )}{\pi_{\theta_{old}}(o_{i,t}|q,o_{i,<t})}$, $\epsilon$ and $\beta$ are hyperparameters, the KL term is defined as
\begin{equation}
\begin{aligned}
\mathbb{D}_{KL} = \frac{\pi_{ref}(o_{i,t}|q,o_{i,<t})}{\pi_\theta(o_{i,t}|q,o_{i,<t})} - \log \frac{\pi_{ref}(o_{i,t}|q,o_{i,<t})}{\pi_\theta(o_{i,t}|q,o_{i,<t})} - 1,
\end{aligned}
\end{equation}

and the advantage $A_i$ is computed using a group of rewards $\{r_1, r_2, ..., r_G\}$ with GRAE:
\begin{equation}
\begin{aligned}
A_i = \frac{r_i - \text{mean}(\{r_1, r_2, ..., r_G\})}{\text{std}(\{r_1, r_2, ..., r_G\})}.
\end{aligned}
\end{equation}

\section{Implicit Advantage Symmetry in GRPO}

First, we demonstrate that GRPO's policy gradient optimization can be formulated as \textbf{a reweighting variant of Supervised Fine-Tuning (SFT)}. Specifically, the gradient update can be formulated as follows (proof in~\cref{sec:proof eq4}):
\begin{equation}
\begin{aligned}
\nabla_\theta \mathcal{J}_{\text{GRPO}}(\theta) = \mathbb{E}_{q \sim \mathcal{Q}, \{o_i\}_{i=1}^G \sim \pi_{\theta_{\text{old}}}(\cdot|q)} \underbrace{\frac{1}{G}\sum_{i=1}^G\left[
\underbrace{\frac{1}{|\mathbf{o}_i|} \sum_{t=1}^{|\mathbf{o}_i|} \underbrace{\rho_{i,t} A_{i,t}}_{\text{weight}} \underbrace{\nabla_\theta \log \pi_\theta(o_{i,t} \mid q, \mathbf{o}_{i,<t})}_{\text{SFT}}}_{\text{Group Level}}
\right]}_{\text{Sample Level}} . 
\end{aligned}
\label{eq:pgo}
\end{equation}
In \cref{eq:pgo}, $\rho_{i,t}$ is relatively stable with the clip operation, so the dominant part of reweighting is $A_{i,t}$. In general, the advantage of GRPO is shared across the entire sequence, meaning that all tokens within a single response correspond to the same advantage value, i.e.,\(A_{i,1}=A_{i,2}=\dots=A_{i,t}=A_{i}\). We then examine this reweighting mechanism from the group level and the sample level.

\textbf{Advantage Symmetry in Group Level:} In RLVR, a trajectory is considered correct provided that the final answer extracted from the response aligns with the ground truth. Given a query, we partition the sampled within-group responses into correct (positive) and incorrect (negative) trajectories.  Crucially, we observe that the aggregate weights of policy updates attributed to correct trajectories are equivalent to those attributed to incorrect ones (proof in~\cref{sec:proof eq5}). This property can be expressed as:

\begin{equation}
\begin{aligned}
\sum_{i\in G_{pos}}|A_{i}|=\sum_{i\in G_{neg}}|A_{i}|
\end{aligned}
\label{eq:5}
\end{equation}
where $G_{pos}$ and $G_{neg}$ represent the sets of positive and negative trajectories, satisfying the conditions $G_{pos} \cap G_{neg} = \emptyset$ and $G_{pos} \cup G_{neg} = G$ (the sampled response set). This property is subsequently evaluated within the behavior space.

\textbf{Theorem 1} (Behavior-Space Logit Update) 
\textit{Assume that the set of all possible behaviors is $B=\{b_{i}\}_{i=1}^{N}$, consisting of sampled set $G$ and unsampled set $U$. Given the intragroup advantage sum $C = \sum_{b_i \in G} A_{i}$, the behavior-level logit update of \(b_i\) under gradient ascent can be expressed as:}
\begin{equation}
\Delta h_{b_i}
=\eta\left[\mathbb I(b_i\in G)A_{b_i}-C\pi_{b_i}\right].
\end{equation}
\textit{where $\eta$ denotes the learning rate and $\pi_{b_i}$ is the model's current sampling probability in behavior space for trajectory $b_i$ (proof in~\cref{sec:proof eq6}).}

In standard GRAE, advantage normalization enforces the zero-sum property $C=0$. Therefore, sampled correct trajectories are directly reinforced, while sampled incorrect trajectories are directly suppressed. In contrast, an unsampled trajectory receives no direct advantage-driven update in the behavior-space abstraction. Moreover, for a parameterized nonlinear policy, shared parameters still indirectly change the probability of an unsampled trajectory as our parameter-space analysis in~\cref{sec:parameterized-theorem1} shows that such changes arise only through gradient alignment with sampled trajectories and contain no reward signal specific to the unsampled trajectory.

\textbf{Advantage Symmetry in Sample Level:} At the sample level, the overall learning contribution can be captured by the sum of absolute advantages, which represents the total weights of one query. To evaluate the relative contributions across queries of varying complexities, we introduce the sample success rate, denoted as $p$ for a group of size $|G|$, as a proxy for task difficulty; specifically, a higher $p$ indicates a lower level of difficulty. By leveraging this metric, we can formally quantify the relationship between sample difficulty and the corresponding update magnitude.


\textbf{Theorem 2} (Update Magnitude with respect to Sample Difficulty). 
\textit{Consider a group $G$ of trajectories for one query (sample) with binary rewards ${\{r_i\}}_{i=1}^G$, the sum of absolute advantages over the group under Group Relative Advantage Estimation (GRAE) can be derived as:}
\begin{equation}
\begin{aligned}
\sum_{i \in G} |A_i| = 2 |G| \sqrt{p(1-p)} .
\end{aligned}
\label{eq:s-sample}
\end{equation}
\textit{where $p=\sum_{i \in G}|r_i|/|G|$ denotes the empirical success probability of the corresponding sample within group $G$, and the proof can be found in~\cref{sec:Proof of Theorem 2}.}

Theorem~2 reveals an intrinsic difficulty bias in GRAE: samples of intermediate difficulty ($p=0.5$) dominate policy updates. As illustrated in \cref{fig:sample-magnitude}, the symmetry of $\sqrt{p(1-p)}$ assigns identical weights to easy ($p=0.75$) and hard ($p=0.25$) samples. This static approach, however, ignores the dynamic distributional shift during training, where simple instances increasingly dominate the sample pool. Such symmetry causes the model to overfit on trivial data at the expense of challenging scenarios, failing to satisfy the non-stationary demands of the evolving learning process.

\vspace{-0.5em}
\section{Deconstructing the Implicit Advantage Symmetry of GRAE}
\label{Deconstructing the Implicit Advantage Symmetry of GRAE}

\subsection{Experimental Setup}
\vspace{-5pt}

\begin{table*}[!h] 
    \centering
    \caption{\textbf{Summary of Controlled Experiments.} We design two experiments to investigate the impact of symmetry at the group and sample levels. $\beta=10$ and $\gamma=0.5$ are scaling constants.}
    \label{tab:symmetry_experiments}

    \begin{subtable}{0.42\textwidth} 
        \centering
        \caption{\textbf{Exp I: Breaking Group-Level Symmetry.}}
        \label{tab:exp_group}
        \renewcommand{\arraystretch}{1.0}
        \resizebox{\linewidth}{!}{%
            \begin{tabular}{l l c }
                \toprule
                \textbf{Variant} & \textbf{Formulation} ($A^*$) & $\sum_{i}^{G} A_i^*$ \\
                \midrule
                GRPO (Control Group) & $A_{\text{pos}}^* = A_{\text{pos}}$ & $=0$ \\
                Positive-Dominant & $A_{\text{pos}}^* = \beta \cdot A_{\text{pos}}$ & $>0$\\
                Negative-Dominant & $A_{\text{pos}}^* = A_{\text{pos}} / \beta$ & $<0$\\
                \bottomrule
            \end{tabular}%
        }
    \end{subtable}
    \hfill 
    \begin{subtable}{0.48\textwidth} 
        \centering
        \caption{\textbf{Exp II: Breaking Sample-Level Symmetry.}}
        \label{tab:exp_sample}
        \renewcommand{\arraystretch}{1.0}
        \resizebox{\linewidth}{!}{%
            \begin{tabular}{l l c }
                \toprule
                \textbf{Variant} & \textbf{Formulation} ($A^*$) & $\sum_{i}^{G} |A_i^*|$ \\
                \midrule
                GRPO (Control Group) & $A_{i}^* = A_{i}$ & $2|G|\sqrt{p(1-p)}$  \\
                Hard-Focused & $A_{i}^* = \gamma \cdot A_{i}/\sqrt{p}$ & $|G|\sqrt{1-p}$  \\
                Easy-Focused & $A_{i}^* = \gamma \cdot A_{i}/\sqrt{1-p}$ & $|G|\sqrt{p}$  \\
                \bottomrule
            \end{tabular}%
        }
    \end{subtable}
\end{table*}

To systematically investigate the impact of advantage symmetry on reasoning performance, we design two sets of ablation experiments as shown in ~\cref{tab:symmetry_experiments}:

\noindent \textbf{Exp I: Breaking Group-Level Symmetry.} 
This experiment isolates the
effect of violating the group-level symmetry by asymmetrically scaling the advantages of correct trajectories. To achieve this, we introduce a scaling coefficient $\beta=10$ to disrupt the zero-sum equilibrium ($\sum A_i = 0$). Denoting the original positive advantage as $A_{\text{pos}}$, we design two variants:
\begin{enumerate}[leftmargin=2em, nosep]
    \item \textit{Positive-Dominant Group:} We scale up positive advantages ($A_{\text{pos}}^* = \beta \cdot A_{\text{pos}}$). 
    \item \textit{Negative-Dominant Group:} We scale down positive advantages ($A_{\text{pos}}^* = A_{\text{pos}} / \beta$).
\end{enumerate}
The original GRPO serves as the control group, representing the symmetric equilibrium.

\noindent \textbf{Exp II: Breaking Sample-Level Symmetry.} 
This experiment investigates difficulty focus during training by modifying advantage magnitudes using the sampling success rate $p$. Let $A_i$ be the original advantage of GRPO. We define two curriculum variants:
\begin{enumerate}[leftmargin=2em, nosep]
    \item \textit{Hard-Focused Group:} We shift the learning focus toward harder queries by rescaling the advantage with the factor $1/\sqrt{p}$ ($A_{i}^* = \gamma \cdot A_{i}/\sqrt{p}$).
    \item \textit{Easy-Focused Group:} Conversely, we shift the focus to simple queries by scaling with the factor $1/\sqrt{1-p}$ ($A_{i}^* = \gamma \cdot A_{i}/\sqrt{1-p}$). 
\end{enumerate}
Here, $\gamma=0.5$ is a normalization constant to ensure that the theoretical maximum value remains consistent with the control group (standard GRPO). Note that no extra rescaling is conducted when the success rate is 0 or 1, as the GRPO advantage is zero in these cases.

\textbf{Models and Training Setup:} We conducte experiments on math-specialized \texttt{Qwen2.5-Math-7B}~\cite{yang2024qwen25mathtechnicalreportmathematical} and the general-purpose \texttt{Llama-3.2-3B-Instruct}~\cite{dubey2024llama}. All models are trained on the MATH dataset~\cite{hendrycks2021measuring} using \texttt{verl} framework~\cite{sheng2025hybridflow}. Key configurations include a batch size of 1,024, $G=8$ rollouts per query, and a learning rate of $1\times 10^{-6}$. We evaluate model performance across MATH, AMC23, and the high-difficulty AIME 2025 benchmarks using an unbiased Pass@$k$ estimator, consistent with established literature~\cite{hochlehnert2025sober,chen2021evaluating,chen2025pass,zhu2025surprising}. Other training details are available in \cref{Hyperparameter Settings}.

\subsection{Experiments on Breaking Group Level Symmetry}
\label{Rethinking Symmetry in Group Level}

    \begin{figure*}[htbp] 
    \centering  
    \begin{subfigure}{0.4\textwidth}  
        \centering
        \includegraphics[width=\textwidth]{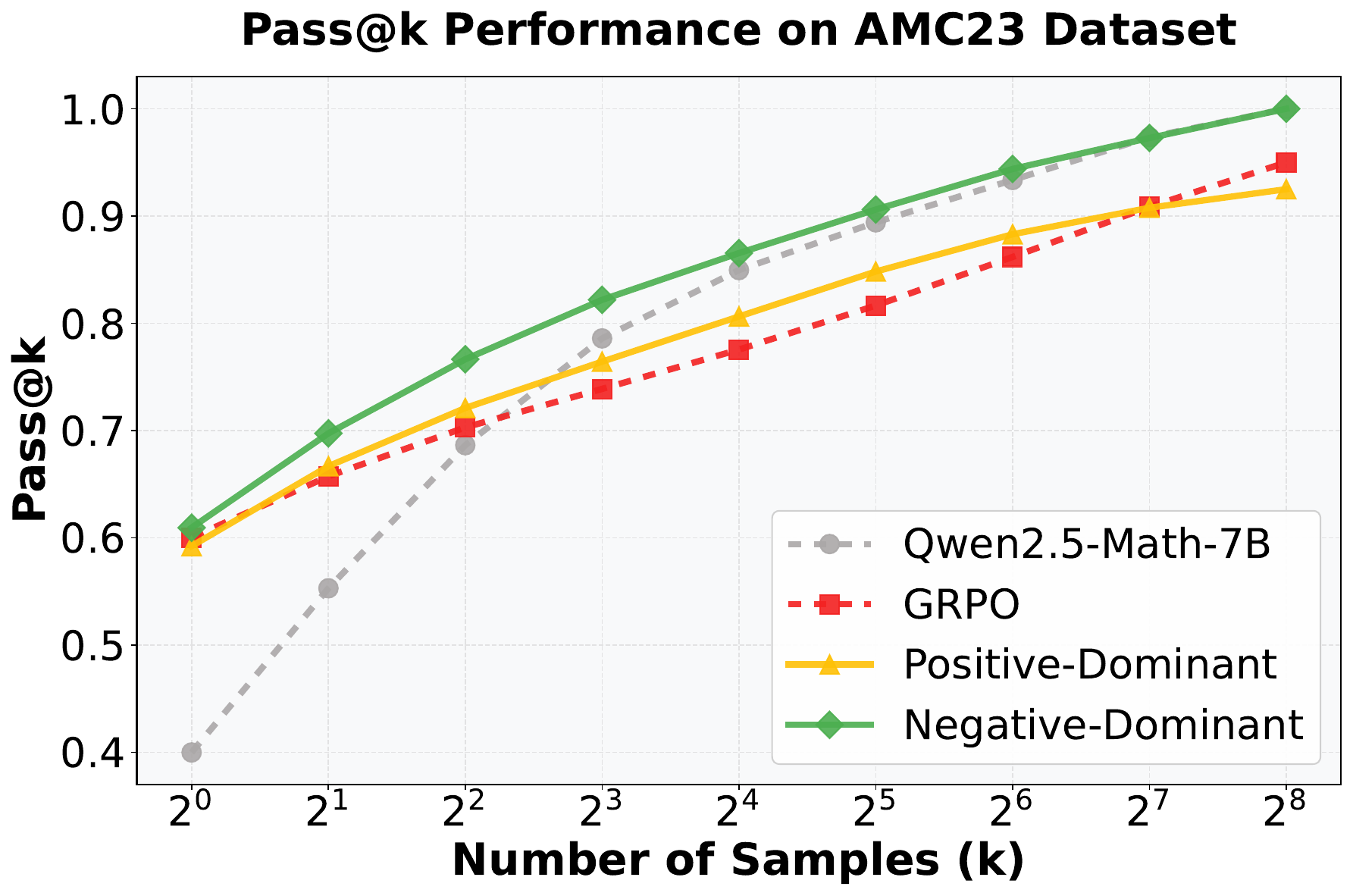}
        \label{subfig:math}
    \end{subfigure}
    \vspace{-10pt}
    \begin{subfigure}{0.4\textwidth} 
        \centering
        \includegraphics[width=\textwidth]{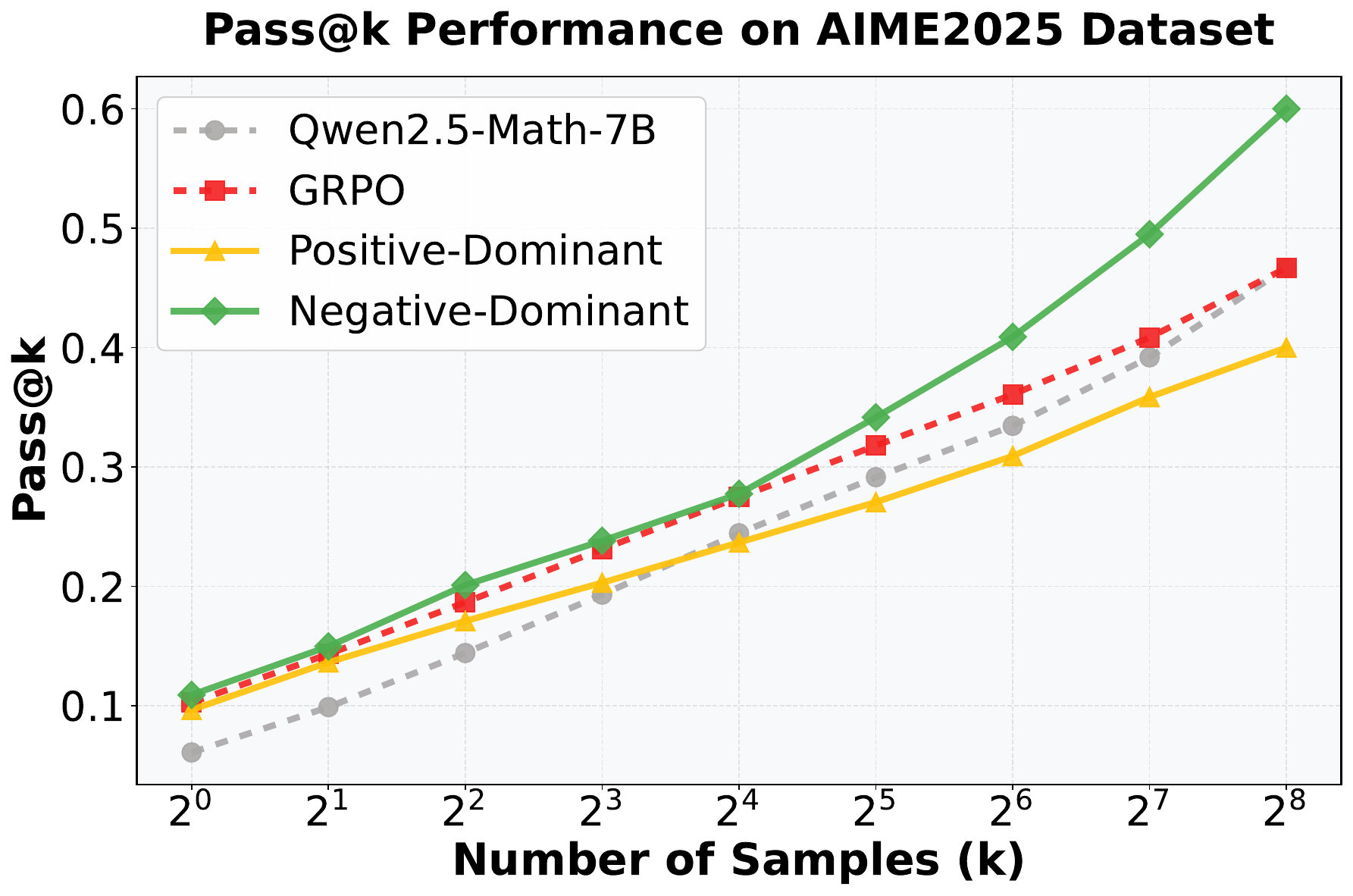}
        \label{subfig:aime25}
    \end{subfigure}
    \vspace{-10pt}
    \begin{subfigure}{0.4\textwidth} 
        \centering
        \includegraphics[width=\textwidth]{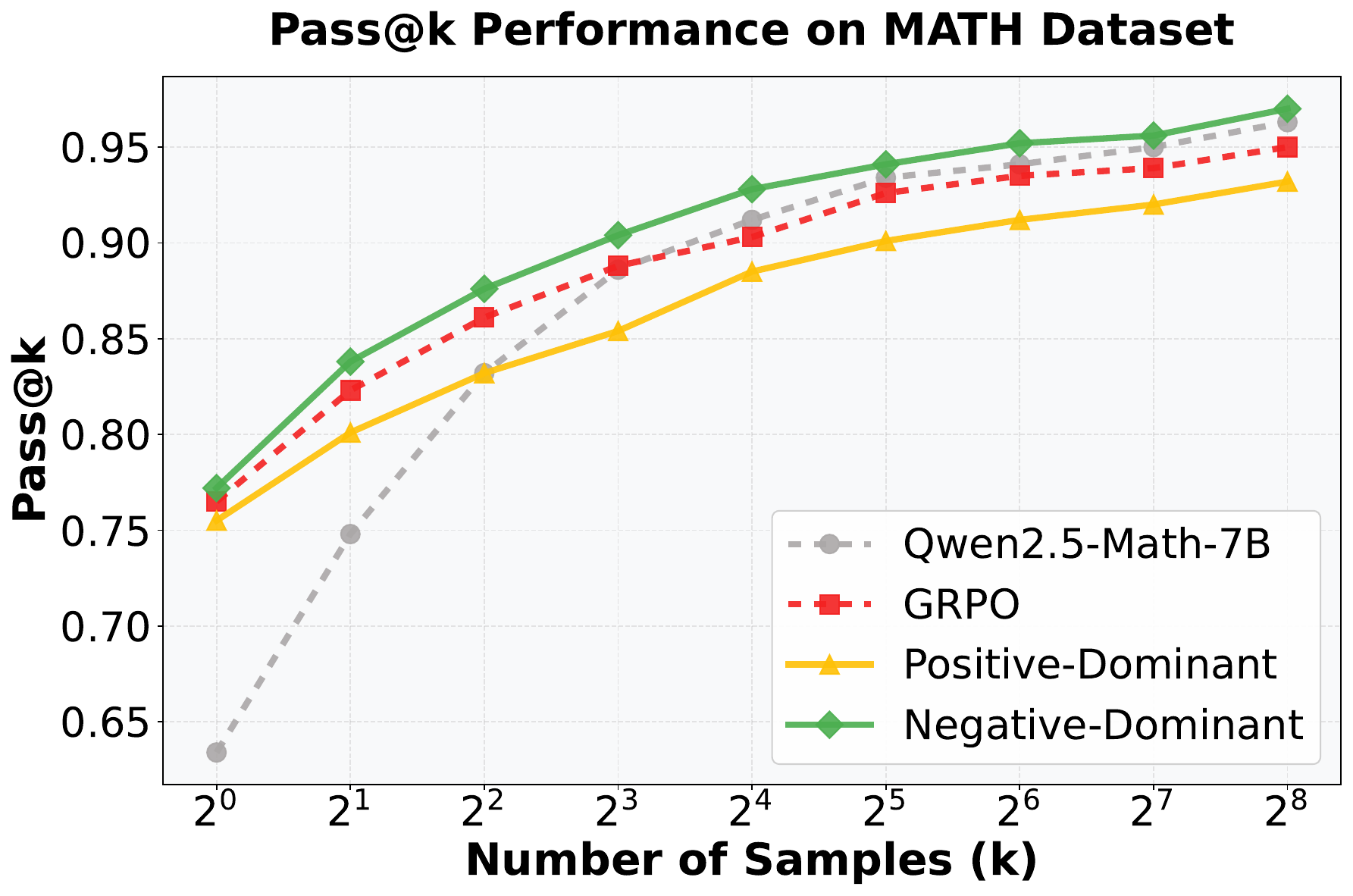}
    \end{subfigure}
    \begin{subfigure}{0.4\textwidth} 
        \centering
        \includegraphics[width=\textwidth]{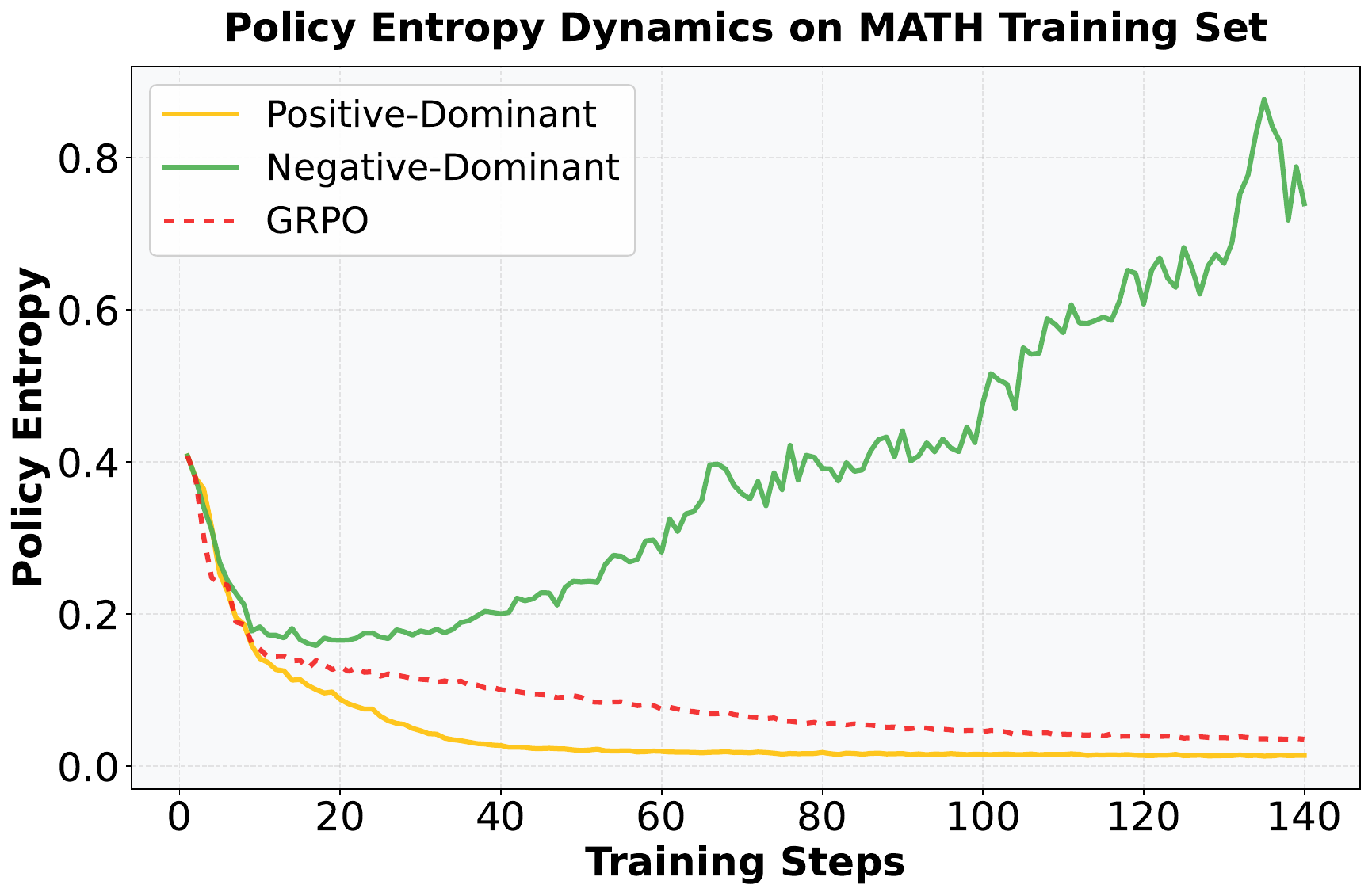}
    \end{subfigure}
    \caption{\textbf{Experimental results on breaking group-level symmetry using \texttt{Qwen2.5-Math-7B}.} We amplify (Positive-Dominant) or suppress (Negative-Dominant) the advantages of correct trajectories to compare their performance with those of GRPO and the base model.The performance is evaluated using Pass@$k$. The last figure illustrates the entropy dynamics on the training set.}
    \label{fig:exp1}
\end{figure*}

\textbf{GRPO enhances efficiency without expanding reasoning boundaries.} As shown in ~\cref{fig:exp1}, while GRPO significantly improves Pass@1 across all datasets, these gains erode at higher sampling budgets. Notably, on the AMC23 and MATH datasets, GRPO’s performance at Pass@256 falls below that of the base model. This suggests that while GRPO improves the sampling probability of correct paths, it fails to discover novel solutions that lie outside the base model's original sampling support. In other words, GRPO does not fundamentally expand the intrinsic reasoning boundaries of the base model. This observation aligns with findings in recent literature~\cite{yue2025does,yao2025debate}.

\textbf{Over-emphasizing correct paths triggers entropy collapse.} In contrast to GRPO, amplifying the weight of positive trajectories fails to yield performance improvements. In particular, on the AIME2025 and MATH datasets, the Positive-Dominant group significantly underperforms other methods at larger sampling budgets ($k>16$). To investigate the underlying cause, we find that the policy entropy measured on the training set for the Positive-Dominant group exhibits the most precipitous decline, as illustrated in~\cref{fig:exp1}. This phenomenon indicates that overemphasis on correct trajectories reduces sampling diversity by excessively sharpening the output distribution, leading to entropy collapse and thus impairing the model’s reasoning capability.

\textbf{Down-weighting the correct path improves performance but risks instability.} Notably, down-weighting positive trajectories consistently outperforms vanilla GRPO in Pass@$k$ metrics across all datasets. Even at a high sampling budget ($k=256$), this Negative-Dominant approach maintains parity with the base model, suggesting it effectively mitigates the capability boundary shrinkage inherent in GRPO. As shown in~\cref{fig:exp1}, this improvement correlates with a sustained increase in policy entropy, indicating enhanced exploration. Importantly, this gain is not merely a byproduct of generic entropy inflation.  As demonstrated in our entropy-matched control experiments (\cref{sec:Comparison Experiments with Entropy Promoting Techniques}), naively promoting entropy degrades performance and induces repetitive outputs. Rather, what matters is how exploration is promoted at the trajectory level: inherently breaking advantage symmetry prevents indiscriminate broad support diffusion and fosters meaningful diversity. However, in practice, the Negative-Dominant strategy can trigger training instability (demonstrated in~\cref{Training Collapse of Negative-Dominant Group}) when overconfident incorrect trajectories displace correct ones (discussed in \cref{sec:the logits of ndg}). Consequently, a dynamic adjustment mechanism is essential to balance exploration diversity with training stability as training progresses.

\begin{tcolorbox}[takeawaybox]
  \textbf{Takeaway 1:}  As discussed in ~\cref{Rethinking Symmetry in Group Level}, while down-weighting positive advantage incentivizes the model to explore unsampled correct trajectories to improve the reasoning ability, it also introduces the risk of training collapse in the later stages of reinforcement learning.
\end{tcolorbox}

\subsection{Experiments on Breaking Sample Level Symmetry}
\label{Rethinking Symmetry in Sample Level}

\begin{figure*}[htbp] 
    \centering  
    \begin{subfigure}{0.4\textwidth}  
        \centering
        \includegraphics[width=\textwidth]{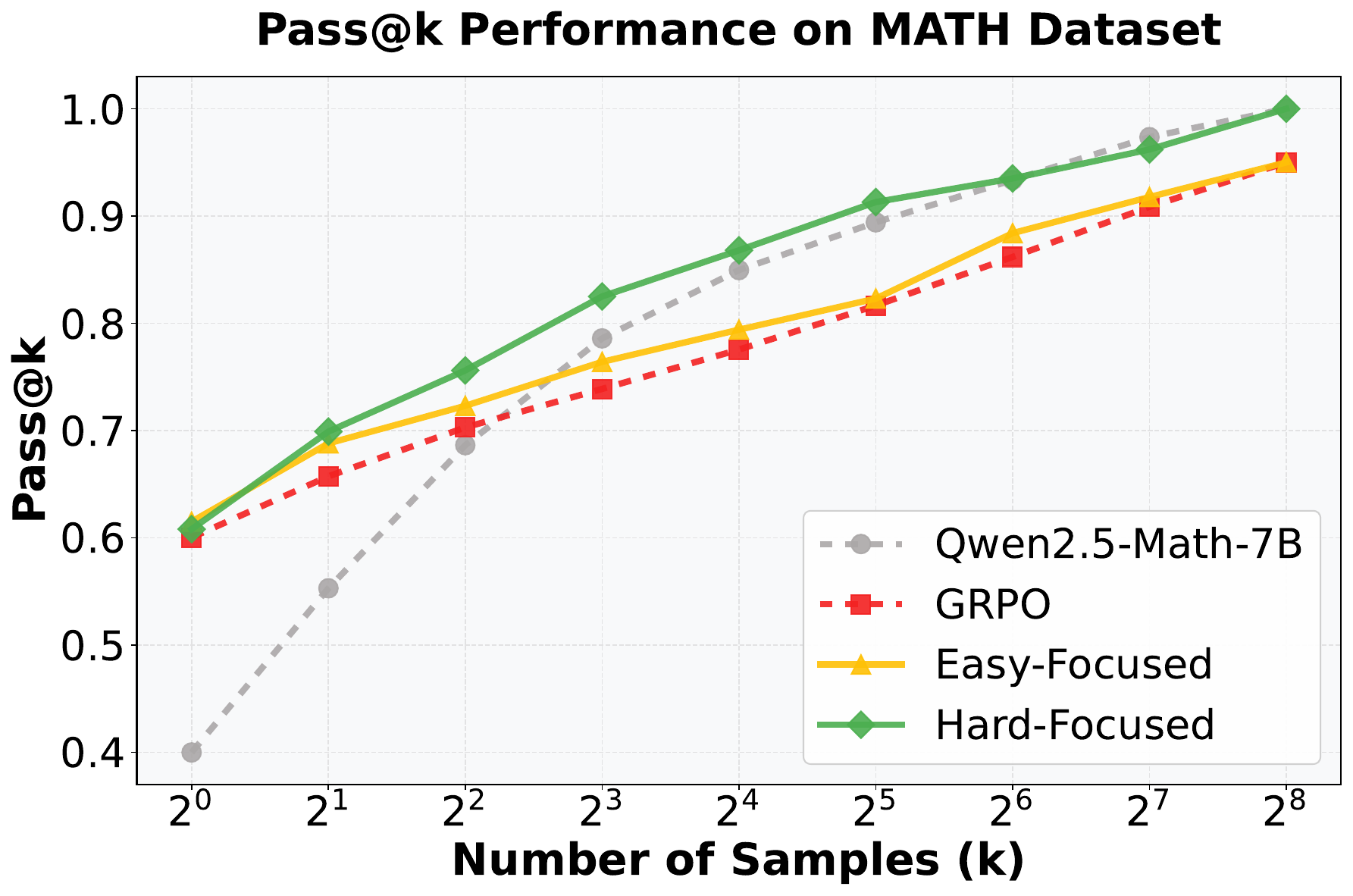}
        \label{subfig:amcs}
    \end{subfigure}
    \vspace{-10pt}
    \begin{subfigure}{0.4\textwidth} 
        \centering
        \includegraphics[width=\textwidth]{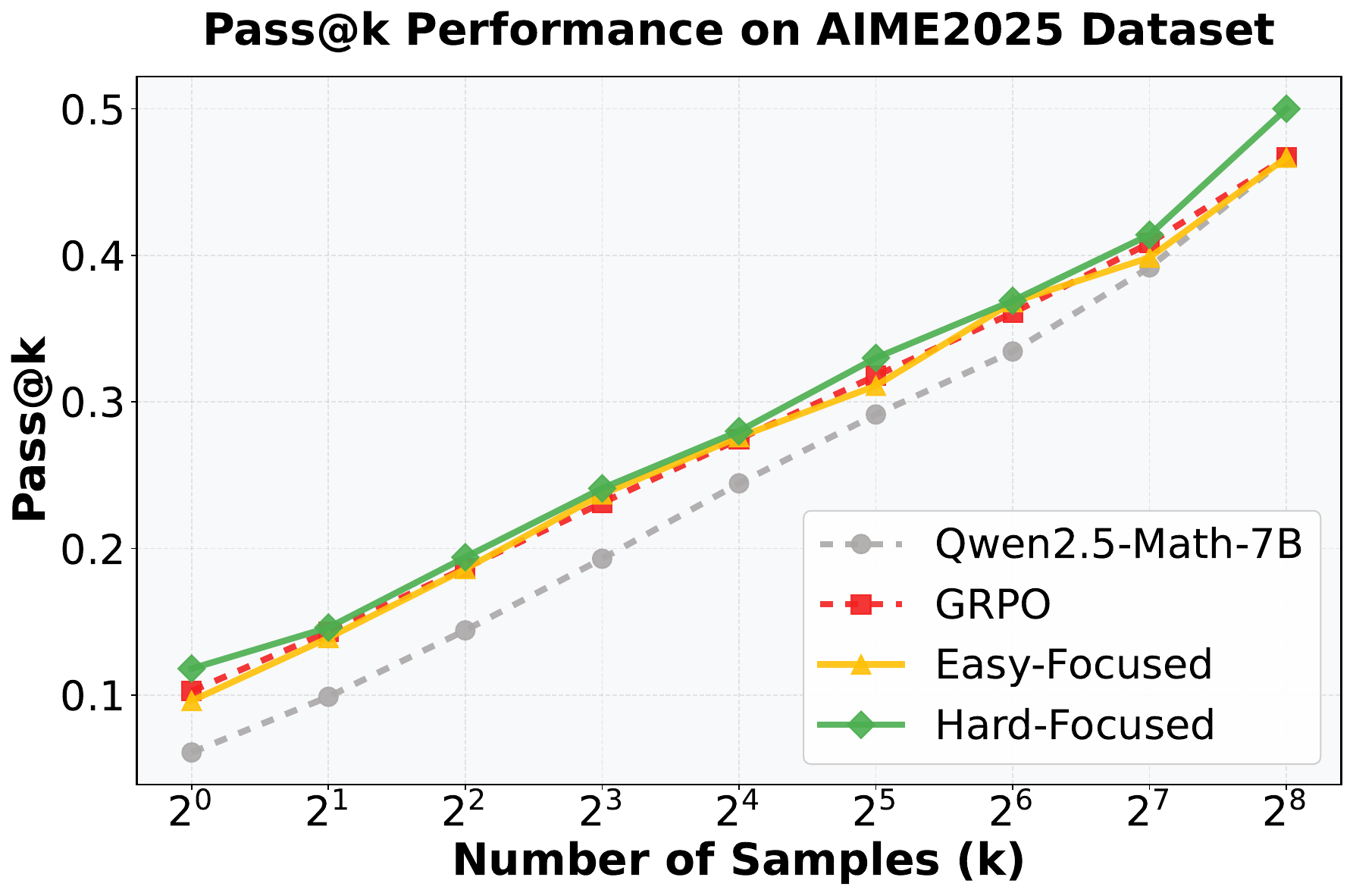}
        \label{subfig:aime25s}
    \end{subfigure}
    \vspace{-10pt}
    \begin{subfigure}{0.4\textwidth} 
        \centering
        \includegraphics[width=\textwidth]{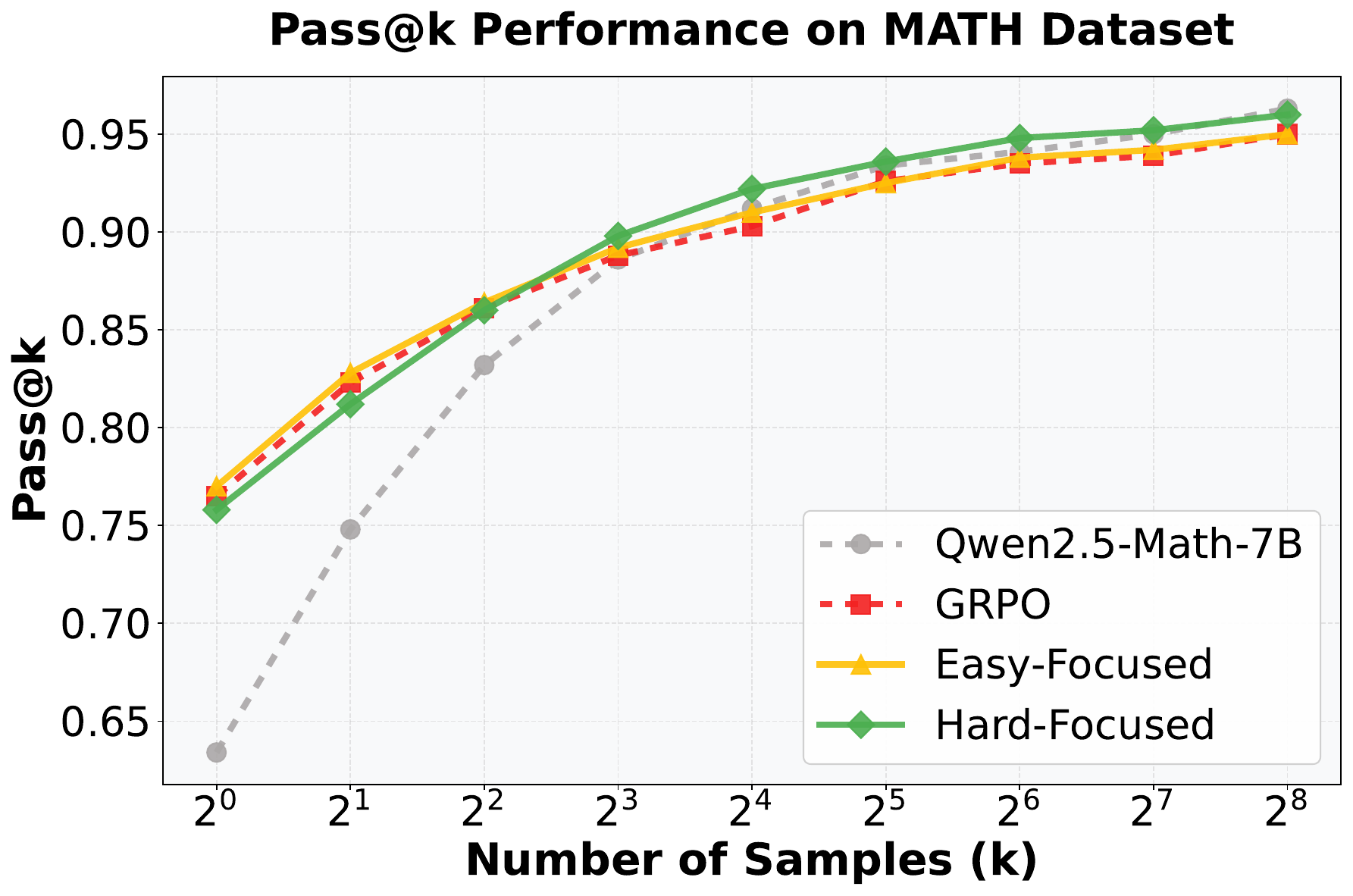}
        \label{subfig:maths}
    \end{subfigure}
    \begin{subfigure}{0.4\textwidth} 
        \centering
        \includegraphics[width=\textwidth]{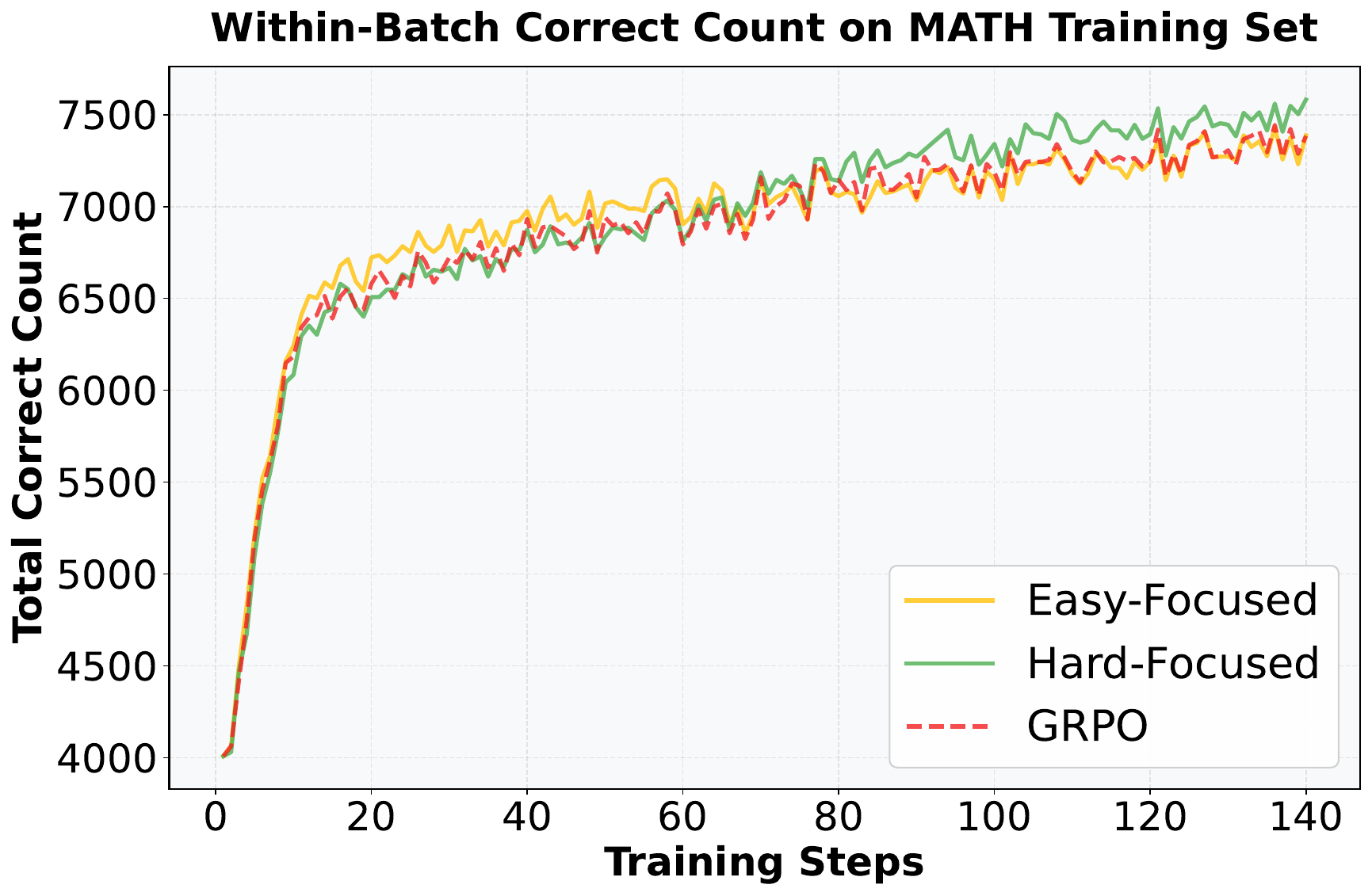}
        \label{subfig:correctcnt}
    \end{subfigure}
    \caption{\textbf{Experimental results on breaking sample-level symmetry using \texttt{Qwen2.5-Math-7B}.} We rescale the advantages to shift the learning focus toward harder queries (Hard-Focused) or easier queries (Easy-Focused) to compare their performance with those of GRPO and the base model.The performance is evaluated using Pass@$k$. The last figure illustrates the within-batch count of correct sampled responses on the training set.}
    \label{fig:exp2}
    \vspace{-10pt}
\end{figure*}

\textbf{Static difficulty reweighting is not universally optimal and depends on test difficulty.} As shown in ~\cref{fig:exp2}, none of the methods yielded a uniform advantage across all datasets. More precisely, their relative efficacy varies by benchmark: hard-focused methods achieve peak results on the challenging AIME2025, whereas easy-focused approaches marginally outperform others on simpler datasets such as AMC23 and MATH at Pass@$1$. These findings indicate the optimal difficulty weighting is evaluation-dependent, challenging the prevailing trend of focusing solely on hard samples throughout training~\cite{zhang2025grpo,pikus2025hard,guan2025emit}.

\textbf{Prioritizing simple samples in early stages and hard samples later yields better learning efficiency gains.} As direct evaluation fails to identify a universally optimal strategy, we further investigate the training dynamics. As shown in~\cref{fig:exp2}, the Easy-Focused strategy accelerates most rapidly in early phases, which suggests that prioritizing simpler tasks promotes the learning of basic formatting rules and core reasoning patterns. Conversely, the Hard-Focused strategy excels in later stages. This shift suggests that once model capability reaches relative saturation, a transition toward difficult samples becomes essential to further elevate the performance ceiling. Unlike prior curriculum RL methods that manually impose fixed "easy-to-hard" schedules based on predefined rankings~\cite{chen2025self,fosterlilo} or focus solely on medium-difficulty or difficult samples~\cite{zhang2025learning,bae2025online}, our findings reveal a non-stationary learning pattern that yields better learning efficiency gains.

\begin{tcolorbox}[takeawaybox]
  \textbf{Takeaway 2:} Our findings in~\cref{Rethinking Symmetry in Sample Level} suggest that static difficulty reweighting is insufficient, as optimal sample utility is phase-dependent. While easy samples facilitate more effective learning in the early stage, a strategic transition toward hard samples becomes useful as performance begins to plateau.
\end{tcolorbox}

\subsection{A Unified Perspective on Existing RL Methods}

The proposed \textbf{advantage symmetry} provides a unified perspective for connecting existing GRPO-based methods that were originally developed from different motivations. From this perspective, negative-reinforcement and Pass@$k$-oriented approaches~\citep{zhu2025surprising,yao2025debate,nan2025ngrpo,li2024turning,wang2024learning} can be interpreted as implicitly breaking \textbf{group-level symmetry} to promote exploration, whereas methods that emphasize high-entropy tokens or difficult samples~\citep{cheng2025reasoning,hao2025rethinking,zhang2025edge,deng2025decomposing,park2025deepvideo,zhang2025grpo,shen2025dast,zhang2025learning,bae2025online} mainly break \textbf{sample-level symmetry} by reallocating learning signals toward uncertain or hard regions. 

Consequently, this unified formulation enables controlled interventions along the two symmetry axes and reveals findings beyond existing fixed asymmetric strategies. At the group level, stronger exploration induced by suppressing positive advantages is beneficial but can become unstable when maintained excessively; at the sample level, a fixed preference for hard samples is not always optimal, and the preferred difficulty shifts from easier queries in early training to harder ones in later stages. These observations suggest that the two forms of symmetry should be \textbf{jointly and dynamically broken according to the training state}, which directly motivates the design of A-GRAE\footnote{More related work is discussed in detail in Section~A.}.


\vspace{-5pt}
\section{Asymmetric GRAE}
\vspace{-5pt}

\begin{table*}[t]
\centering
\caption{Pass@$k$ results on MATH, AIME 2025 and AMC23 with \texttt{Qwen2.5-Math-7B}. \best{Bold} and \second{underlined} numbers denote the best and second-best results for each $k$. Specifically, We conduct 16 runs on AIME2025 and AMC23, and 4 runs on the remaining benchmarks. }
\label{tab:math-results}
\small
\setlength{\tabcolsep}{3pt} 
\renewcommand{\arraystretch}{0.95} 

\newcommand{\var}[1]{\hspace{1.5pt}{\scriptsize$\pm#1$}}

\resizebox{0.9\textwidth}{!}{
\begin{tabular}{l cccccccccc}
\toprule
\best{Method} & \multicolumn{10}{c}{\best{Pass@$k$}}  \\
\cmidrule(lr){2-11}
$k$ & 1 & 2 & 4 & 8 & 16 & 32 & 64 & 128 & 256 & \best{Avg.} \\
\midrule
\multicolumn{11}{c}{\best{MATH}} \\
Base Model & 63.4\var{0.4} & 74.8\var{0.5} & 83.2\var{0.6} & 88.6\var{0.7} & 91.2\var{0.7} & 93.4\var{0.6} & 94.1\var{0.5} & 95.0\var{0.4} & 96.3\var{0.3} & 86.7 \\
SEC & 77.6\var{0.5} & 82.5\var{0.6} & 86.0\var{0.6} & 88.9\var{0.4} & 90.2\var{0.5} & 92.2\var{0.6} & 92.6\var{0.5} & 93.3\var{0.3} & 94.7\var{0.3} & 88.7 \\
GRPO-LEAD & 77.8\var{0.3} & 83.0\var{0.4} & 86.5\var{0.5} & 89.2\var{0.5} & 90.5\var{0.5} & 92.3\var{0.5} & 92.8\var{0.4} & 93.6\var{0.4} & 95.0\var{0.3} & 89.0 \\
P@kT & 76.2\var{0.6} & 82.2\var{0.8} & 86.9\var{0.6} & 89.6\var{0.4} & 91.4\var{0.3} & 93.3\var{0.4} & 94.0\var{0.5} & 96.2\var{0.4} & 96.2\var{0.4} & 89.3 \\
W-REINFORCE & 76.6\var{0.5} & 82.8\var{0.6} & 87.1\var{0.7} & 90.2\var{0.8} & 92.4\var{0.8} & 94.1\var{0.7} & \second{95.3\var{0.6}} & \second{96.1\var{0.5}} & \second{96.7\var{0.4}} & 90.1 \\
GRPO & 76.5\var{0.3} & 82.3\var{0.4} & 86.1\var{0.5} & 88.8\var{0.5} & 90.3\var{0.5} & 92.6\var{0.5} & 93.5\var{0.4} & 93.9\var{0.4} & 95.0\var{0.3} & 88.8 \\
\rowcolor{aliceblue} GRPO + A-GRAE & \second{78.3\var{0.2}} & \second{85.0\var{0.3}} & \second{89.2\var{0.4}} & \best{91.0\var{0.4}} & \second{92.5\var{0.4}} & \second{94.6\var{0.4}} & 95.0\var{0.3} & 95.5\var{0.3} & 96.5\var{0.2} & \second{90.8} \\
DAPO & 75.0\var{0.4} & 79.8\var{0.5} & 85.0\var{0.6} & 88.4\var{0.6} & 89.8\var{0.6} & 92.0\var{0.6} & 92.8\var{0.5} & 93.4\var{0.4} & 94.3\var{0.4} & 87.8 \\
\rowcolor{aliceblue} DAPO + A-GRAE & 76.9\var{0.3} & 82.6\var{0.4} & 86.5\var{0.5} & 89.2\var{0.5} & 90.6\var{0.5} & 92.8\var{0.5} & 93.6\var{0.4} & 94.2\var{0.4} & 95.3\var{0.3} & 89.1 \\
Dr.GRPO & 77.2\var{0.4} & 82.5\var{0.5} & 87.4\var{0.6} & 89.6\var{0.6} & 91.0\var{0.6} & 92.8\var{0.5} & 93.6\var{0.5} & 94.3\var{0.4} & 95.0\var{0.4} & 89.3 \\
\rowcolor{aliceblue} Dr.GRPO + A-GRAE & \best{78.6\var{0.3}} & \best{86.2\var{0.3}} & \best{89.8\var{0.4}} & \second{90.6\var{0.4}} & \best{92.8\var{0.4}} & \best{95.0\var{0.4}} & \best{95.4\var{0.3}} & \best{96.0\var{0.3}} & \best{96.9\var{0.2}} & \best{91.3} \\
\midrule
\multicolumn{11}{c}{\best{AIME 2025}} \\
Base Model & 6.1\var{0.7} & 9.9\var{0.9} & 14.4\var{1.1} & 19.3\var{1.3} & 24.4\var{1.5} & 29.1\var{1.6} & 33.4\var{1.6} & 39.2\var{1.5} & 46.7\var{1.3} & 24.7 \\
SEC & 11.3\var{0.8} & 15.2\var{1.0} & 19.5\var{0.9} & 23.8\var{1.2} & 28.3\var{1.5} & 32.1\var{1.8} & 36.4\var{1.6} & 41.0\var{1.3} & 49.0\var{1.2} & 28.5 \\
GRPO-LEAD & 11.0\var{0.5} & 14.8\var{0.7} & 19.2\var{0.9} & 23.4\var{1.1} & 27.8\var{1.3} & 32.0\var{1.4} & 36.5\var{1.5} & 41.4\var{1.4} & 47.3\var{1.3} & 28.2 \\
P@kT & 10.8\var{0.5} & 15.6\var{0.6} & 20.6\var{1.2} & 25.0\var{1.4} & 29.8\var{1.3} & 34.5\var{1.5} & 40.3\var{1.4} & 47.6\var{1.6} & 56.4\var{1.5} & \second{31.2} \\
W-REINFORCE & 10.6\var{0.7} & 15.3\var{0.9} & 20.0\var{1.1} & 24.7\var{1.3} & \second{29.7\var{1.5}} & \second{34.6\var{1.6}} & \second{40.5\var{1.7}} & 47.8\var{1.6} & \second{56.7\var{1.4}} & 31.1 \\
GRPO & 10.3\var{0.5} & 14.3\var{0.7} & 18.7\var{0.9} & 23.1\var{1.1} & 27.5\var{1.3} & 31.8\var{1.4} & 36.1\var{1.5} & 40.8\var{1.4} & 46.7\var{1.3} & 27.7 \\
\rowcolor{aliceblue} GRPO + A-GRAE & 11.3\var{0.4} & 15.6\var{0.6} & 19.8\var{0.8} & 24.7\var{1.0} & 28.6\var{1.2} & 34.1\var{1.3} & 39.2\var{1.3} & 47.8\var{1.2} & \second{56.7\var{1.1}} & 30.9 \\
DAPO & \second{12.0\var{0.6}} & 16.1\var{0.8} & \second{21.3\var{1.0}} & \second{25.2\var{1.2}} & 29.4\var{1.4} & 33.2\var{1.5} & 38.5\var{1.5} & 45.4\var{1.4} & 53.3\var{1.2} & 30.5 \\
\rowcolor{aliceblue} DAPO + A-GRAE & \best{13.3\var{0.4}} & \best{18.4\var{0.6}} & \best{23.0\var{0.8}} & \best{26.3\var{1.0}} & \best{30.0\var{1.2}} & \best{35.1\var{1.3}} & \best{41.1\var{1.3}} & \best{48.7\var{1.2}} & \best{60.0\var{1.0}} & \best{32.9} \\
Dr.GRPO & 11.0\var{0.6} & 14.8\var{0.8} & 19.3\var{1.0} & 24.3\var{1.2} & 28.8\var{1.4} & 33.0\var{1.5} & 37.1\var{1.5} & 41.2\var{1.4} & 46.7\var{1.3} & 28.5 \\
\rowcolor{aliceblue} Dr.GRPO + A-GRAE & 11.8\var{0.4} & \second{16.2\var{0.6}} & 19.8\var{0.8} & 25.0\var{1.0} & 29.3\var{1.2} & 34.8\var{1.3} & 37.9\var{1.3} & \second{48.0\var{1.2}} & \second{56.7\var{1.1}} & 31.1 \\
\midrule
\multicolumn{11}{c}{\best{AMC23}} \\
Base Model & 40.6\var{0.5} & 55.3\var{0.6} & 68.6\var{0.8} & 78.6\var{0.9} & 85.0\var{0.9} & 89.4\var{0.8} & 93.4\var{0.7} & \best{97.3\var{0.5}} & \best{100.0\var{0.0}} & 78.7 \\
SEC & 61.8\var{0.5} & 67.5\var{0.8} & 72.8\var{0.7} & 77.5\var{1.0} & 81.0\var{0.8} & 84.6\var{0.6} & 87.8\var{0.6} & 90.2\var{0.6} & 92.1\var{0.5} & 79.5 \\
GRPO-LEAD & 62.3\var{0.4} & 68.0\var{0.5} & 73.3\var{0.6} & 77.8\var{0.7} & 81.5\var{0.8} & 85.0\var{0.7} & 88.2\var{0.6} & 90.3\var{0.5} & 92.3\var{0.4} & 79.9 \\
P@kT & 61.3\var{0.5} & 68.8\var{0.8} & 76.0\var{0.8} & 82.3\var{1.0} & 87.3\var{0.8} & 91.6\var{0.7} & 94.8\var{0.6} & 96.7\var{0.5} & 97.2\var{0.6} & 84.0 \\
W-REINFORCE & 62.0\var{0.6} & 70.0\var{0.7} & 77.0\var{0.8} & 83.1\var{0.9} & 87.8\var{0.9} & 91.8\var{0.8} & 95.0\var{0.7} & \second{97.0\var{0.5}} & \second{97.5\var{0.4}} & 84.6 \\
GRPO & 60.2\var{0.4} & 66.7\var{0.5} & 72.1\var{0.6} & 76.4\var{0.7} & 80.6\var{0.7} & 84.8\var{0.7} & 88.3\var{0.6} & 90.8\var{0.5} & 92.5\var{0.4} & 79.2 \\
\rowcolor{aliceblue} GRPO + A-GRAE & \second{62.6\var{0.3}} & 70.0\var{0.4} & 77.5\var{0.5} & 83.7\var{0.6} & 88.2\var{0.7} & 92.0\var{0.7} & \second{95.1\var{0.6}} & 96.8\var{0.4} & \second{97.5\var{0.3}} & 84.8 \\
DAPO & 62.0\var{0.4} & \second{70.3\var{0.5}} & 77.2\var{0.6} & 83.1\var{0.7} & 87.8\var{0.8} & 91.4\var{0.8} & 94.0\var{0.7} & 96.1\var{0.5} & \second{97.5\var{0.4}} & 84.4 \\
\rowcolor{aliceblue} DAPO + A-GRAE & \best{63.3\var{0.3}} & \best{72.5\var{0.4}} & \best{80.5\var{0.5}} & \best{86.7\var{0.6}} & \best{90.2\var{0.7}} & \best{92.9\var{0.7}} & 95.0\var{0.6} & \second{97.0\var{0.4}} & \best{100.0\var{0.0}} & \best{86.5} \\
Dr.GRPO & 60.7\var{0.5} & 69.8\var{0.6} & 75.6\var{0.7} & 82.6\var{0.8} & 87.8\var{0.8} & 90.9\var{0.8} & 93.2\var{0.7} & 94.6\var{0.6} & 95.0\var{0.5} & 83.4 \\
\rowcolor{aliceblue} Dr.GRPO + A-GRAE & 62.8\var{0.3} & 71.6\var{0.4} & \second{78.2\var{0.5}} & \second{84.0\var{0.6}} & \second{89.6\var{0.7}} & \second{92.3\var{0.7}} & \best{95.2\var{0.6}} & 95.9\var{0.5} & \best{100.0\var{0.0}} & \second{85.5} \\
\bottomrule
\end{tabular}}
\vspace{-10pt}
\end{table*}

\subsection{Method}
\vspace{-5pt}


Our prior analysis reveals that the advantage symmetry inherent in GRPO undermines model exploration and difficulty adaptation. To address these limitations, we propose the Asymmetric Group Relative Advantage Estimation (A-GRAE) framework to dynamically modulate exploration incentives and sample-difficulty focus. To implement this, a metric is required to quantify  training state; accordingly, we introduce the batch-wise mean reward as a proxy indicator:
\begin{equation}
\begin{aligned}
\omega_s = \frac{\sum_{i=1}^{B}r_i}{B},
\end{aligned}
\end{equation}
where $B$ denotes the total number of trajectories in the batch, $\omega_s$ denotes the mean reward of the current step, where a higher value implies stronger model proficiency. Then we introduce a dynamic attention shift at the sample level, transitioning from easy to hard samples as training progresses:
\begin{equation}
\begin{aligned}
A_i = \frac{\omega_s}{2}\cdot\frac{r_i - \text{mean}(\{r_1, r_2, ..., r_G\})}{\text{std}(\{r_1, r_2, ..., r_G\})\cdot\sqrt{p}}+\frac{1-\omega_s}{2}\cdot\frac{r_i - \text{mean}(\{r_1, r_2, ..., r_G\})}{\text{std}(\{r_1, r_2, ..., r_G\})\cdot\sqrt{1-p}},
\end{aligned}
\end{equation}
where $p$ denotes the sampling success rate for a given query. Note that rescaling is omitted when $p\in\{0,1\}$, as the standard advantage equals 0. As the model evolves with increasing sampling success rate, the weight assigned to the hard-focused component (first term) progressively increases, while the corresponding weight for the easy-focused component (second term) diminishes. This mechanism facilitates adaptive trade-offs, dynamically shifting the training focus toward harder questions as the model’s proficiency improves.

At the group level, we propose an attenuation suppression strategy for correct-response advantages, which encourages exploration in the early stage while preserving stability in the later phase:
\begin{equation}
\label{eq:attn}
\begin{aligned}
A_i^* = \begin{cases} 
A_i*\min(1, \frac{\omega_s}{\alpha}), & \text{if } A_i > 0 \\
A_i, & \text{if } A_i \le 0 
\end{cases}
\end{aligned}
\end{equation}
Here, $\alpha>0$ is a scaling parameter. Once the refined advantages are computed, they can be seamlessly incorporated into the GRPO function, as defined in~\cref{eq:grpo}, or other GRPO variants for optimization.

\vspace{-5pt}
\subsection{Experimental Setup}
\vspace{-5pt}

\begin{table}[htbp]
\centering
\caption{Results on multi-modal benchmarks with \texttt{Qwen2.5-VL-3B-Instruct}. We conduct 4 runs on each benchmark. \best{Bold} and \second{underlined} numbers denote the best and second-best results.}
\label{tab:mm_results}
\setlength{\tabcolsep}{5pt} 
\renewcommand{\arraystretch}{0.95} 

\newcommand{\std}[1]{\hspace{1.5pt}{\scriptsize$\pm#1$}}

\resizebox{0.4\textwidth}{!}{ 
\begin{tabular}{l c cc} 
\toprule
& \best{ID Domain} & \multicolumn{2}{c}{\best{OOD Domain}} \\
\cmidrule(lr){2-2} \cmidrule(lr){3-4}
\best{Method} & Geo3K & MathVision & Mathverse \\
\midrule
\rowcolor{gray!5} \multicolumn{4}{l}{\textit{Task A: General Mathematical Reasoning}} \\
Base Model & 27.8\std{1.0} & 20.8\std{0.8} & 31.6\std{0.7} \\
GRPO & 43.5\std{0.8} & 23.4\std{0.7} & 35.2\std{0.9} \\
\rowcolor{aliceblue} GRPO + A-GRAE & \second{45.7}\std{0.6} & 24.0\std{0.4} & 36.8\std{0.5} \\
DAPO & 44.7\std{0.9} & 23.8\std{0.6} & 35.9\std{0.8} \\
\rowcolor{aliceblue} DAPO + A-GRAE & 45.9\std{0.4} & \second{24.3}\std{0.5} & \second{37.5}\std{0.8} \\
Dr.GRPO & 44.9\std{0.7} & 24.2\std{0.8} & 36.5\std{0.7} \\
\rowcolor{aliceblue}Dr.GRPO + A-GRAE & \best{46.8}\std{0.5} & \best{25.6}\std{0.6} & \best{38.4}\std{0.6} \\
\midrule
\rowcolor{gray!5} \multicolumn{4}{l}{\textit{Task B: Medical Imaging Reasoning}} \\
& MRI300 & CT300 & Xray300 \\
\cmidrule(lr){2-2} \cmidrule(lr){3-4}
Base Model & 35.6\std{1.6} & 42.5\std{2.1} & 42.0\std{1.4} \\
GRPO & 87.2\std{1.2} & 71.7\std{0.9} & 63.2\std{1.1} \\
\rowcolor{aliceblue} GRPO + A-GRAE & \second{88.2}\std{0.8} & \second{73.1}\std{0.7} & 71.3\std{1.0} \\
DAPO & 84.3\std{1.1} & 71.6\std{1.0} & 63.1\std{0.9} \\
\rowcolor{aliceblue} DAPO + A-GRAE & 87.0\std{1.0} & 72.3\std{0.8} & \second{71.6}\std{0.7} \\
Dr.GRPO & 87.8\std{0.9} & 72.4\std{0.8} & 69.5\std{0.7} \\
\rowcolor{aliceblue}Dr.GRPO + A-GRAE & \best{89.0}\std{0.6} & \best{73.6}\std{0.7} & \best{72.0}\std{0.5} \\
\bottomrule
\end{tabular}}
\vspace{-10pt}
\end{table}

 We validate our method across seven benchmarks, including text-only mathematical tasks (AIME2025, AMC23, MATH~\cite{hendrycks2021measuring}) with \texttt{Qwen2.5-Math-7B} and \texttt{DeepSeek-R1-7B} (\cref{sec:deepseek}) and multimodal tasks in mathematics (Geo3K~\cite{lu2021inter}, MathVision~\cite{wang2024measuring}, MathVerse~\cite{zhang2024mathverse}) and medicine (HuatuoGPT-Vision~\cite{chen2024huatuogpt}) with \texttt{Qwen2.5-VL-3B-Instruct}. Our experiments evaluate GRPO variants (GRPO, DAPO, Dr.GRPO) alongside representative baselines. These include pass@k optimization methods targeting exploration efficiency (W-REINFORCE~\cite{zhu2025surprising}, p@kT~\cite{chen2025pass}) and curriculum learning approaches for difficulty adaptation (GRPO-LEAD~\cite{zhang2025grpo}, SEC~\cite{chen2025self}). We report Pass@{k} on the text datasets and  Pass@{1} on the multimodal datasets since most of them are multiple-choice questions. For details  please refer to~\cref{Hyperparameter Settings}.

\vspace{-5pt}
\subsection{Main Results}
\label{seq:main-results}
\vspace{-5pt}

\textbf{Consistent improvements in Pass@$1$ and Pass@$k$.} As illustrated in~\cref{tab:math-results}, our proposed method yields substantial performance gains when integrated with various GRPO variants, which proves the effectiveness of our proposed method (the statistical significance tests can be found in~\cref{sec:detailed_p_value}). Notably, existing baselines often exhibit a trade-off between accuracy and diversity. For instance, curriculum learning approaches SEC and GRPO-LEAD primarily improves Pass@1 but shows limited gains in Pass@$k$ categories. Conversely, pass@k optimization methods W-REINFORCE and P@kT emphasize Pass@$k$ performance while achieving relatively modest improvements in greedy accuracy. In contrast, our A-GRAE realizes consistent gains across both metrics, which proves the superiority of our method.



\textbf{Universal applicability to multi-modal domains.} To evaluate the versatility of A-GRAE, we extended our empirical analysis to the multimodal domain, with the results summarized in~\cref{tab:mm_results}. The results reveal that A-GRAE not only delivers significant improvements in the in-distribution (ID) domain but also yields substantial gains in out-of-distribution (OOD) scenarios. These findings provide strong evidence that our approach effectively enhances sampling efficiency while simultaneously preserving the model's generalization capabilities. In summary, the efficacy of A-GRAE across diverse domains and tasks underscores its universal applicability as a robust framework.

\vspace{-5pt}
\subsection{Further Analysis}
\vspace{-5pt}

\textbf{Ablation studies.} To validate the contribution of each component in A-GRAE, we perform a series of ablation studies, and the results are demonstrated in~\cref{tab:ablation}. Firstly, sample-level asymmetry primarily bolsters Pass@$1$ performance, consistently attaining second-best results across all three benchmarks. Conversely, the group-level asymmetric mechanism exerts a more pronounced impact on enhancing Pass@$k$ metrics, suggesting that this module effectively fosters generative diversity. Ultimately, the full framework yields the most significant  gains across all evaluation metrics, demonstrating that the constituent modules of our methodology are mutually complementary. 

\textbf{Training dynamics.} Figure~\ref{fig:dynamics} illustrates the training dynamics, specifically entropy and greedy decoding accuracy on the MATH dataset. While GRPO suffers from a continuous entropy decline on the training set, A-GRAE maintains a sustained plateau after an initial drop, successfully averting entropy collapse. On the test set, A-GRAE follows a "rise-then-fall" entropy trajectory, characteristic of a balanced exploration-exploitation paradigm. This dynamic enables A-GRAE to significantly outpace GRPO in greedy decoding accuracy during the latter training stages, confirming its superior capability in fostering sustained learning.

\vspace{-10pt}
\section{Conclusion}
\vspace{-10pt}
In this paper, we rethink the mechanics of Group Relative Advantage Estimation (GRAE) within the GRPO framework. Our theoretical and empirical analyses reveal a previously overlooked ``advantage symmetry" in standard GRPO, and we demonstrate that this symmetry restricts exploration and fails to adapt its difficulty focus during the learning process. To address these limitations, we propose A-GRAE, a novel mechanism that dynamically modulates exploration incentives and prioritizes samples based on their evolving utility. Extensive evaluations across seven benchmarks demonstrate the effectiveness and superiority of A-GRAE.



\bibliography{ref}
\bibliographystyle{plainnat}


\appendix

\section{Related Works}
\label{sec:relatedworks}

\subsection{Advantage Estimation in Reinforcement Learning.}
 In Proximal Policy Optimization (PPO), advantage estimation relies on the Generalized Advantage Estimation (GAE~\citep{schulman2015high}) framework. Nevertheless, the critic model required by GAE introduces non-trivial computational overhead. To mitigate this cost, GRPO~\citep{shao2024deepseekmath} and REINFORCE++~\citep{hu2025reinforce++} replace the critic with lightweight baselines (batch-averaged rewards) or group-relative rewards for advantage computation. Building on this, several methods have pointed out and addressed the limitations of GRPO’s group relative advantage estimation. Dr.GRPO~\citep{liu2025understanding} enhances token efficiency by removing length and standard deviation normalization terms, while DAPO~\citep{yu2025dapo} balances the policy gradient loss at the token level to mitigate the insufficient gradient contribution of long sequences in long CoT scenarios. However, the advantage symmetry phenomenon inherent in GRPO's advantage estimation remains under-explored, with the existing literature lacking an in-depth analysis of this property.

\subsection{GRPO Variants for Exploration Enhancement.}
Recent studies have observed that RLVR may improve greedy accuracy while reducing solution diversity at larger sampling budgets~\citep{yue2025does}.
To mitigate this issue, several methods~\citep{dang2025assessing,he2025rewarding,ma2025learning,gandhi2025cognitive} explicitly encourage exploration through refined optimized objectives. 
Negative-reinforcement approaches~\citep{zhu2025surprising,nan2025ngrpo} suppress incorrect trajectories or reduce the relative reinforcement of correct ones, thereby redistributing probability mass toward alternative solutions. 
Pass@$k$-oriented methods~\citep{chen2025pass,walder2025passk} instead directly optimize the collective utility of multiple sampled responses, encouraging diverse solutions. 

From the perspective of advantage symmetry, these methods can be interpreted as different ways of breaking \textbf{group-level symmetry} to increase exploration. 
However, they primarily focus on how to strengthen exploration, while overlooking the training instability brought by the symmetry breaking. 
Our controlled experiments reveal that stronger group-level asymmetry is not always preferable: although down-weighting positive advantages substantially improves Pass@$k$, excessive or persistent negative dominance can cause training collapse. 
This finding motivates our attenuation strategy of A-GRAE, which promotes exploration when needed while preserving learning stability.

\subsection{GRPO Variants for Difficulty Adaptation.}
Another line of work improves GRPO by adapting the learning signal to problem difficulty. 
Since queries of different difficulty contribute unequally to policy optimization, difficulty-aware methods assign larger weights to challenging or informative samples~\citep{zhang2025grpo,shen2025dast,zhang2025learning,bae2025online,dai2026harder}, while curriculum-based approaches adjust the training distribution according to some designed data selection principles~\citep{zhou2025codapo,zhang2025clpo}. 
These strategies generally improve overall accuracy by preventing optimization from being dominated by already-solved or less informative queries.

However, existing approaches generally impose a fixed difficulty preference or course paradigm, while how the optimal preference changes throughout training remains less explored. 
Our controlled experiments instead reveal a stage-dependent pattern: easier queries yield larger gains in the early stage, whereas harder queries become more beneficial as the model have reached a plateau in the latter stage. Accordingly, A-GRAE dynamically shifts its learning focus from easier to harder queries rather than maintaining a fixed difficulty bias.

\section{Supplementary Proof}

\subsection{Proof of Equation 4}
\label{sec:proof eq4}

\paragraph{Proof.} We begin by reviewing the objective function of GRPO under the notation used in our formulation.
\begin{equation}
\begin{aligned}
\mathcal{J}_{\text{GRPO}}(\theta) = &\mathbb{E}_{q \sim \mathcal{Q}, \{o_i\}_{i=1}^G \sim \pi_{\theta_{\text{old}}}(\cdot|q)} \\
&\Bigg[ \frac{1}{G} \sum_{i=1}^G \frac{1}{|o_i|} \sum_{t=1}^{|o_i|} \Big\{  \min \Big( \rho_{i,t}(\theta) A_{i,t}, \text{clip}(\rho_{i,t}(\theta), 1\pm\epsilon) A_{i,t} \Big) \Big\}  - \beta \mathbb{D}_{\text{KL}}(\pi_\theta || \pi_{\text{ref}}) \Bigg] ,
\end{aligned}
\end{equation}

where $\rho_{i,t}(\theta) = \frac{\pi_\theta(o_{i,t} \mid q, o_{i,<t})}{\pi_{\theta_{\text{old}}}(o_{i,t} \mid q, o_{i,<t})}$, and $\beta$ is the coefficient for the KL divergence term.

To better understand the model's learning dynamics under the verifiable reward setting and to derive the specific reweighting formulation, we omit the regularization components firstly (e.g., KL term \& clipping operation) and focus on the core policy optimization objective:

\begin{equation}
\mathcal{J}_{\text{GRPO}}(\theta) = \mathbb{E}_{q \sim \mathcal{Q}, \{o_i\}_{i=1}^G \sim \pi_{\theta_{\text{old}}}(\cdot|q)} \left[ \frac{1}{G} \sum_{i=1}^G \frac{1}{|o_i|} \sum_{t=1}^{|o_i|} \rho_{i,t}(\theta) A_{i,t} \right] ,
\end{equation}

Taking the gradient with respect to $\theta$:

\begin{align}
&\nabla_\theta \mathcal{J}_{\text{GRPO}}(\theta) = \mathbb{E}_{q \sim \mathcal{Q}, \{o_i\}_{i=1}^G \sim \pi_{\theta_{\text{old}}}(\cdot|q)} \left[ \frac{1}{G} \sum_{i=1}^G \frac{1}{|o_i|} \sum_{t=1}^{|o_i|} A_{i,t} \nabla_\theta \rho_{i,t}(\theta) \right] \notag \\
&= \mathbb{E}_{q \sim \mathcal{Q}, \{o_i\}_{i=1}^G \sim \pi_{\theta_{\text{old}}}(\cdot|q)} \left[ \frac{1}{G} \sum_{i=1}^G \frac{1}{|o_i|} \sum_{t=1}^{|o_i|} A_{i,t} \frac{\nabla_\theta \pi_\theta(o_{i,t} \mid q, o_{i,<t})}{\pi_{\theta_{\text{old}}}(o_{i,t} \mid q, o_{i,<t})} \right] \notag \\
&= \mathbb{E}_{q \sim \mathcal{Q}, \{o_i\}_{i=1}^G \sim \pi_{\theta_{\text{old}}}(\cdot|q)} \left[ \frac{1}{G} \sum_{i=1}^G \frac{1}{|o_i|} \sum_{t=1}^{|o_i|} A_{i,t} \frac{\pi_\theta(o_{i,t} \mid q, o_{i,<t})}{\pi_{\theta_{\text{old}}}(o_{i,t} \mid q, o_{i,<t})} \nabla_\theta \log \pi_\theta(o_{i,t} \mid q, o_{i,<t}) \right] \notag \\
&= \mathbb{E}_{q \sim \mathcal{Q}, \{o_i\}_{i=1}^G \sim \pi_{\theta_{\text{old}}}(\cdot|q)} \left[ \frac{1}{G} \sum_{i=1}^G \frac{1}{|o_i|} \sum_{t=1}^{|o_i|} \underbrace{\rho_{i,t}(\theta) A_{i,t}}_{\text{weight}} \underbrace{\nabla_\theta \log \pi_\theta(o_{i,t} \mid q, o_{i,<t})}_{\text{SFT}} \right] .
\end{align}
We complete the proof of~\cref{eq:pgo}. As the weight equal to 1, GRPO degrade to the standard SFT function, thus it can be formulated as a reweighting variant of Supervised Fine-Tuning.

Then we consider the full GRPO objective function including the clipping mechanism and the KL regularization term:
\begin{equation}
\begin{aligned}
\mathcal{J}_{\text{GRPO}}(\theta) = \mathbb{E}_{q \sim \mathcal{Q}, \{o_i\}_{i=1}^G \sim \pi_{\theta_{\text{old}}}} 
\Bigg[ \frac{1}{G} \sum_{i=1}^G \frac{1}{|o_i|} \sum_{t=1}^{|o_i|} \mathcal{L}_{i,t}^{\text{CLIP}}(\theta) - \beta \mathbb{D}_{\text{KL}}(\pi_\theta || \pi_{\text{ref}}) \Bigg] ,
\end{aligned}
\end{equation}
where the clipped surrogate objective is defined as:
\begin{equation}
\mathcal{L}_{i,t}^{\text{CLIP}}(\theta) = \min \left( \rho_{i,t}(\theta) A_{i,t}, \text{clip}(\rho_{i,t}(\theta), 1-\epsilon, 1+\epsilon) A_{i,t} \right).
\end{equation}

Taking the gradient with respect to $\theta$, we apply the chain rule. For the first term, we introduce an indicator function $\mathbb{I}_{i,t}$ to represent the clipping condition:
\begin{equation}
\nabla_\theta \mathcal{L}_{i,t}^{\text{CLIP}}(\theta) = \mathbb{I}_{i,t} \cdot A_{i,t} \nabla_\theta \rho_{i,t}(\theta) = \mathbb{I}_{i,t} \cdot \rho_{i,t}(\theta) A_{i,t} \nabla_\theta \log \pi_\theta(o_{i,t} \mid q, o_{i,<t}),
\end{equation}
where $\mathbb{I}_{i,t} = 1$ if the ratio $\rho$ is within the trust region or if the update further improves the objective without being clipped, and $\mathbb{I}_{i,t} = 0$ otherwise.

For the KL term, using the standard approximation $\nabla_\theta \mathbb{D}_{\text{KL}}(\pi_\theta || \pi_{\text{ref}}) \approx \nabla_\theta \log \pi_\theta$, the total gradient becomes:
\begin{equation}
\begin{aligned}
\nabla_\theta \mathcal{J}_{\text{GRPO}}(\theta) = \mathbb{E} \Bigg[ \frac{1}{G} \sum_{i=1}^G \frac{1}{|o_i|} \sum_{t=1}^{|o_i|} \Big( \underbrace{\mathbb{I}_{i,t} \cdot \rho_{i,t}(\theta) A_{i,t}}_{\text{Gated Reward Weight}} - \underbrace{\beta \frac{\nabla_\theta \mathbb{D}_{\text{KL}}}{\nabla_\theta \log \pi_\theta}}_{\text{KL Penalty Weight}} \Big) \nabla_\theta \log \pi_\theta(o_{i,t} \mid q, o_{i,<t}) \Bigg].
\end{aligned}
\end{equation}

Defining the effective weight as $W_{i,t} = \mathbb{I}_{i,t} \rho_{i,t}(\theta) A_{i,t} - \beta \psi_{i,t}$, we obtain:
\begin{equation}
\nabla_\theta \mathcal{J}_{\text{GRPO}}(\theta) = \mathbb{E} \left[ \frac{1}{G} \sum_{i=1}^G \frac{1}{|o_i|} \sum_{t=1}^{|o_i|} W_{i,t} \cdot \nabla_\theta \log \pi_\theta(o_{i,t} \mid q, o_{i,<t}) \right].
\end{equation}
This confirms that GRPO, even with all regularization terms, maintains the form of a \textbf{Gated Reweighted SFT}, where the gating is controlled by clipping and the scale is adjusted by the KL penalty.

\subsection{Proof of Equation 5}
\label{sec:proof eq5}

Given a group of sampled responses indexed by $i \in \{1, \dots, G\}$, partitioned into a set of correct trajectories $G_{pos}$ and incorrect trajectories $G_{neg}$. Let the advantage $A_i$ be computed via GRAE as:
\begin{equation}
A_i = \frac{r_i - \mu}{\sigma}, \quad \text{where } \mu = \frac{1}{G}\sum_{j=1}^G r_j, \quad \sigma = \sqrt{\frac{1}{G}\sum_{j=1}^G (r_j - \mu)^2}.
\end{equation}
Then, the sum of absolute advantages for correct trajectories equals that of incorrect trajectories:
\begin{equation}
\sum_{i\in G_{pos}}|A_{i}|=\sum_{i\in G_{neg}}|A_{i}|.
\end{equation}

\paragraph{Proof.} 
First, we examine the sum of the standardized advantages over the entire group. By definition:
\begin{align}
\sum_{i=1}^G A_i &= \sum_{i=1}^G \frac{r_i - \mu}{\sigma} \notag \\
&= \frac{1}{\sigma} \left( \sum_{i=1}^G r_i - \sum_{i=1}^G \mu \right) \notag \\
&= \frac{1}{\sigma} \left( \sum_{i=1}^G r_i - G \cdot \frac{1}{G}\sum_{j=1}^G r_j \right) \notag \\
&= \frac{1}{\sigma} \left( \sum_{i=1}^G r_i - \sum_{i=1}^G r_i \right) = 0.
\end{align}
This confirms the zero-sum property of the standardized advantages. Now, we decompose the summation into the two disjoint sets $G_{pos}$ and $G_{neg}$ (assuming the group contains mixed results, i.e., $G_{pos} \cup G_{neg} = G$):
\begin{equation}
\sum_{i=1}^G A_i = \sum_{i \in G_{pos}} A_i + \sum_{i \in G_{neg}} A_i = 0.
\end{equation}
Rearranging the terms, we obtain:
\begin{equation}
\sum_{i \in G_{pos}} A_i = - \sum_{i \in G_{neg}} A_i.
\end{equation}
In the context of verifiable rewards (e.g., binary outcomes $r \in \{0, 1\}$), correct trajectories receive higher rewards than the group mean ($r_i > \mu$), implying $A_i > 0$ for $i \in G_{pos}$. Conversely, incorrect trajectories receive lower rewards ($r_i < \mu$), implying $A_i < 0$ for $i \in G_{neg}$. We have:

\begin{align}
\sum_{i \in G_{pos}} |A_i| = \sum_{i \in G_{neg}} |A_i|.
\end{align}
Thus, the magnitude of policy updates attributed to correct trajectories is strictly equivalent to that of incorrect ones.

\subsection{Proof of Theorem 1}
\label{sec:proof eq6}

In this section, we first derive Theorem 1 induced by the advantage term without clipping and KL regularization for clarity, and then extend the theorem to the full GRPO objective, confirming that it remains valid in a more general setting.

\textbf{Theorem 1} (Behavior-Space Logit Update) 
\textit{Assume that the set of all possible behaviors is $B=\{b_{i}\}_{i=1}^{N}$, consisting of sampled set $G$ and unsampled set $U$. Given the intragroup advantage sum $C = \sum_{o_i \in G} A_{i}$, the behavior-level logit update of \(b_i\) under gradient ascent can be expressed as:}
\begin{equation}
\Delta h_{b_i}
=\eta\left[\mathbb I(b_i\in G)A_{b_i}-C\pi_{b_i}\right].
\end{equation}
\textit{where $\eta$ denotes the learning rate and $\pi_{b_i}$ is the model's current sampling probability in behavior space for path $b_i$.}

\paragraph{Proof.}
Fix a prompt \(x\). Denote by \(B=\{b_1,b_2,\ldots,b_N\}\) the behavior space, where each \(b_i\) represents a complete trajectory or response. For the current prompt, the sampled group is \(G=\{b^{(1)},b^{(2)},\ldots,b^{(g)}\}\subseteq B\), with group size \(g\), and the unsampled set is \(U=B\setminus G\). In LLM reasoning, \(B\) is combinatorially large while \(g\) is typically small; hence \(U\) contains the overwhelming majority of possible trajectories.

Let \(h=(h_{b_1},h_{b_2},\ldots,h_{b_N})^\top\) be the behavior-level logits. The induced policy over \(B\) is given by the Softmax parameterization
\begin{equation}
    \pi_{b_i}
    =
    \pi_h(b_i\mid x)
    =
    \frac{\exp(h_{b_i})}
    {\sum_{j=1}^{N}\exp(h_{b_j})}.
    \label{eq:behavior-softmax}
\end{equation}
Thus, \(h_{b_i}\) directly controls the probability mass assigned to trajectory \(b_i\): increasing \(h_{b_i}\) raises \(\pi_{b_i}\), while decreasing it lowers \(\pi_{b_i}\). Here we first omit clipping and KL regularization, and consider the simplified policy-gradient objective as
\begin{equation}
    J_{\mathrm{pg}}(h)
    =
    \sum_{b_k\in G}
    \widehat A_{b_k}\log \pi_{b_k},
    \label{eq:simplified-pg-objective}
\end{equation}
where \(\widehat A_{b_k}\) denotes the group-relative advantage of the sampled trajectory \(b_k\). 

For an arbitrary trajectory \(b_i\in B\), the gradient of \(J_{\mathrm{pg}}\) with respect to its logit \(h_{b_i}\) is
\begin{equation}
    \frac{\partial J_{\mathrm{pg}}}{\partial h_{b_i}}
    =
    \sum_{b_k\in G}
    \widehat A_{b_k}
    \frac{\partial \log \pi_{b_k}}{\partial h_{b_i}} .
    \label{eq:gradient-chain-rule}
\end{equation}
By the standard log-softmax derivative,
\begin{equation}
    \frac{\partial \log \pi_{b_k}}{\partial h_{b_i}}
    =
    \mathbb I(b_i=b_k)-\pi_{b_i}.
    \label{eq:log-softmax-derivative}
\end{equation}
Substituting Eq.~\eqref{eq:log-softmax-derivative} into Eq.~\eqref{eq:gradient-chain-rule} yields
\begin{align}
    \frac{\partial J_{\mathrm{pg}}}{\partial h_{b_i}}
    &=
    \sum_{b_k\in G}
    \widehat A_{b_k}
    \big(
        \mathbb I(b_i=b_k)-\pi_{b_i}
    \big) \notag \\
    &=
    \sum_{b_k\in G}
    \widehat A_{b_k}\mathbb I(b_i=b_k)
    -
    \pi_{b_i}
    \sum_{b_k\in G}\widehat A_{b_k}.
    \label{eq:gradient-decomposition}
\end{align}
Assuming the sampled trajectories are distinct, the first term is nonzero only when \(b_i\in G\), and therefore
\[
    \sum_{b_k\in G}
    \widehat A_{b_k}\mathbb I(b_i=b_k)
    =
    \mathbb I(b_i\in G)\widehat A_{b_i}.
\]
Together with the definition of \(C\), we obtain
\begin{equation}
    \frac{\partial J_{\mathrm{pg}}}{\partial h_{b_i}}
    =
    \mathbb I(b_i\in G)\widehat A_{b_i}
    -
    C\pi_{b_i}.
    \label{eq:simplified-gradient-final}
\end{equation}
Applying gradient ascent with learning rate \(\eta>0\), we can obtain:
\begin{equation}
    \Delta h_{b_i}
    =
    \eta
    \left[
        \mathbb I(b_i\in G)\widehat A_{b_i}
        -
        C\pi_{b_i}
    \right].
    \label{eq:simplified-dynamics}
\end{equation}
We complete the proof.

\paragraph{Extension of Theorem 1 considering the clipping and KL terms.} We now refine Theorem 1 by incorporating the clipping mechanism and the
KL regularization term in the complete GRPO objective. For a sampled trajectory
\(b_k\in G\), define the importance ratio
$\rho_{b_k}=
    \frac{\pi_{b_k}}{\pi_{\mathrm{old},b_k}} .
    \label{eq:importance-ratio}$
The clipped surrogate assigns an effective weight to each sampled trajectory:
\begin{equation}
    w_{b_k}
    =
    \rho_{b_k}\widehat A_{b_k}
    \cdot
    \mathbb I
    \left[
    \big(
        \widehat A_{b_k}\ge 0
        \ \wedge\
        \rho_{b_k}\le 1+\epsilon
    \big)
    \vee
    \big(
        \widehat A_{b_k}<0
        \ \wedge\
        \rho_{b_k}\ge 1-\epsilon
    \big)
    \right].
    \label{eq:effective-weight}
\end{equation}
Here \(w_{b_k}\) denotes the actual advantage-driven gradient weight after
importance reweighting and clipping. With this effective weight, the refined Theorem 1 under the
complete GRPO objective are
\begin{equation}
    \Delta h_{b_i}
    =
    \eta
    \left[
        \underbrace{
        \mathbb I(b_i\in G)w_{b_i}
        -
        \pi_{b_i}
        \sum_{b_k\in G}w_{b_k}
        }_{\text{advantage-driven term}}
        +
        \underbrace{
        \beta
        \big(
            \pi_{\mathrm{ref},b_i}-\pi_{b_i}
        \big)
        }_{\text{KL-regularization term}}
    \right].
    \label{eq:complete-behavior-dynamics}
\end{equation}
Therefore, even in the complete formulation, if $b_i \in U$, the term $\mathbb{I}(b_i \in G) w_{b_i}$ vanishes, meaning unsampled correct trajectories still receive no direct positive reinforcement from the environment rewards. They are updated only by the normalization force (the summation of weights) and the KL penalty, which forces them to track the reference distribution rather than exploring novel high-reward regions.

\paragraph{Proof.}
The clipped advantage component of the complete GRPO objective can be written as
\begin{equation}
    J_{\mathrm{adv}}(h)
    =
    \sum_{b_k\in G}
    \min
    \left(
        \rho_{b_k}\widehat A_{b_k},
        \operatorname{clip}
        \big(
            \rho_{b_k},1-\epsilon,1+\epsilon
        \big)
        \widehat A_{b_k}
    \right).
    \label{eq:complete-adv-objective}
\end{equation}
By the definition of \(w_{b_k}\), its logit gradient takes the weighted
log-likelihood form
\begin{equation}
    \frac{\partial J_{\mathrm{adv}}}{\partial h_{b_i}}
    =
    \sum_{b_k\in G}
    w_{b_k}
    \frac{\partial \log \pi_{b_k}}{\partial h_{b_i}} .
    \label{eq:weighted-adv-gradient}
\end{equation}
Using the log-Softmax derivative
\[
    \frac{\partial \log \pi_{b_k}}{\partial h_{b_i}}
    =
    \mathbb I(b_i=b_k)-\pi_{b_i},
\]
we obtain
\begin{align}
    \frac{\partial J_{\mathrm{adv}}}{\partial h_{b_i}}
    &=
    \sum_{b_k\in G}
    w_{b_k}
    \big(
        \mathbb I(b_i=b_k)-\pi_{b_i}
    \big) \notag \\
    &=
    \sum_{b_k\in G}
    w_{b_k}\mathbb I(b_i=b_k)
    -
    \pi_{b_i}
    \sum_{b_k\in G}w_{b_k}.
    \label{eq:weighted-adv-gradient-decomposition}
\end{align}
Assuming sampled trajectories are distinct, the first term simplifies to
\[
    \sum_{b_k\in G}
    w_{b_k}\mathbb I(b_i=b_k)
    =
    \mathbb I(b_i\in G)w_{b_i}.
\]
Therefore,
\begin{equation}
    \frac{\partial J_{\mathrm{adv}}}{\partial h_{b_i}}
    =
    \mathbb I(b_i\in G)w_{b_i}
    -
    \pi_{b_i}
    \sum_{b_k\in G}w_{b_k}.
    \label{eq:weighted-adv-gradient-final}
\end{equation}

It remains to include the KL regularization. In GRPO, the KL penalty pulls the
current policy toward the reference policy \(\pi_{\mathrm{ref}}\). At the
behavior level, this regularization contributes the reference-pullback update
\begin{equation}
    \Delta h_{b_i}^{\mathrm{KL}}
    =
    \eta\beta
    \big(
        \pi_{\mathrm{ref},b_i}-\pi_{b_i}
    \big).
    \label{eq:kl-pullback}
\end{equation}
Combining Eq.~\eqref{eq:weighted-adv-gradient-final} with
Eq.~\eqref{eq:kl-pullback}, and applying a gradient ascent step with learning
rate \(\eta\):
\begin{equation}
    \Delta h_{b_i}
    =
    \eta
    \left[
        \mathbb I(b_i\in G)w_{b_i}
        -
        \pi_{b_i}
        \sum_{b_k\in G}w_{b_k}
        +
        \beta
        \big(
            \pi_{\mathrm{ref},b_i}-\pi_{b_i}
        \big)
    \right],
\end{equation}
Thus the proof is completed.

\textbf{Extension of Theorem 1 to Parameterized Neural Policies}
\label{sec:parameterized-theorem1}.Theorem~1 treats behavior-level logits as independent variables for
analytical clarity. We now show how its interpretation extends to a
nonlinear policy $\pi_\theta$ with shared parameters.

Ignoring clipping and KL regularization for clarity, a gradient-ascent
step under the group-relative policy-gradient objective is
\begin{equation}
\Delta\theta
=
\eta\sum_{b_i\in G}
A_i\nabla_\theta\log\pi_\theta(b_i\mid x).
\end{equation}
For an unsampled trajectory $u\notin G$, a first-order Taylor expansion
gives
\begin{align}
\Delta\log\pi_\theta(u\mid x)
\approx
\nabla_\theta\log\pi_\theta(u\mid x)^\top\Delta\theta 
=
\eta\sum_{b_i\in G}
A_i K_\theta(u,b_i\mid x),
\end{align}
where
\begin{equation}
K_\theta(u,b_i\mid x)
=
\left\langle
\nabla_\theta\log\pi_\theta(u\mid x),
\nabla_\theta\log\pi_\theta(b_i\mid x)
\right\rangle .
\end{equation}

This expression shows that parameter sharing can indeed change the
probability of an unsampled trajectory. However, such a change is
mediated entirely by its gradient alignment with sampled trajectories;
there is no reward term associated with $u$ itself.

To further characterize this indirect update, decompose
\begin{equation}
K_\theta(u,b_i\mid x)
=
\bar K_u+\delta_i,
\end{equation}
where $\bar K_u$ denotes the group-shared alignment component and
$\delta_i$ the sample-specific deviation. Since standard GRAE satisfies
$\sum_{b_i\in G}A_i=0$, we obtain
\begin{equation}
\Delta\log\pi_\theta(u\mid x)
\approx
\eta\sum_{b_i\in G}A_i\delta_i .
\end{equation}
Thus, the common component of parameter sharing cancels exactly.

Furthermore, assuming that, conditional on $u$ being unsampled, the
normalized advantage $A_i$ and the sample-specific alignment deviation
$\delta_i$ are uncorrelated,
\begin{equation}
\mathbb E[A_i\delta_i
\mid x,u\notin G]
=
\operatorname{Cov}
(A_i,\delta_i\mid x,u\notin G)
=0.
\end{equation}
Consequently,
\begin{equation}
\mathbb E
\left[
\Delta\log\pi_\theta(u\mid x)
\mid x,u\notin G
\right]
\approx 0.
\end{equation}

Therefore, shared nonlinear parameters may induce incidental changes
in unsampled trajectories, but standard GRAE provides no systematic
reward-directed update toward an unseen trajectory before it is
sampled. This is the parameter-space counterpart of the behavior-space
result in Theorem~1.

\subsection{Proof of Theorem 2}
\label{sec:Proof of Theorem 2}

\textbf{Theorem 2} (Update Magnitude with respect to Sample Difficulty). 
\textit{Consider a group $G$ of trajectories for one query (sample) with binary rewards ${\{r_i\}}_{i=1}^G$, the sum of absolute advantages over the group under Group Relative Advantage Estimation (GRAE) can be derived as:}
\begin{equation}
\begin{aligned}
\sum_{i \in G} |A_i| = 2 |G| \sqrt{p(1-p)} .
\end{aligned}
\label{eq:s-sample-proof}
\end{equation}
\textit{where $p=\sum_{i \in G}|r_i|/|G|$ denotes the empirical success probability of the corresponding sample within group $G$, and the proof can be found in~\cref{sec:Proof of Theorem 2}.}

\paragraph{Proof.} Given a group $G$ of trajectories with binary rewards $r_i \in \{0, 1\}$, let $p=\sum_{i \in G}|r_i|/|G|$ denote the success probability. The group mean $\mu$ and standard deviation $\sigma$ are:
\begin{equation}
\mu = p, \quad \sigma = \sqrt{p(1-p)}
\end{equation}
The advantage in GRAE is defined as $A_i = (r_i - \mu)/\sigma$. Summing the absolute advantages over the group $G$, we partition the sum into successful ($r_i=1$) and failed ($r_i=0$) subsets:
\begin{equation}
\begin{aligned}
\sum_{i \in G} |A_i| &= \sum_{r_i=1} \left| \frac{1-p}{\sqrt{p(1-p)}} \right| + \sum_{r_i=0} \left| \frac{-p}{\sqrt{p(1-p)}} \right| \\
&= (G \cdot p) \frac{1-p}{\sqrt{p(1-p)}} + (G \cdot (1-p)) \frac{p}{\sqrt{p(1-p)}} \\
&= \frac{2G p (1-p)}{\sqrt{p(1-p)}} \\
&= 2|G|\sqrt{p(1-p)}
\end{aligned}
\end{equation}
This completes the proof, and a visualization of the relationship between update magnitude and sample difficulty is provided below, illustrating the derived curve:

\begin{figure}[htbp]
\centering
\includegraphics[width=0.6\linewidth]{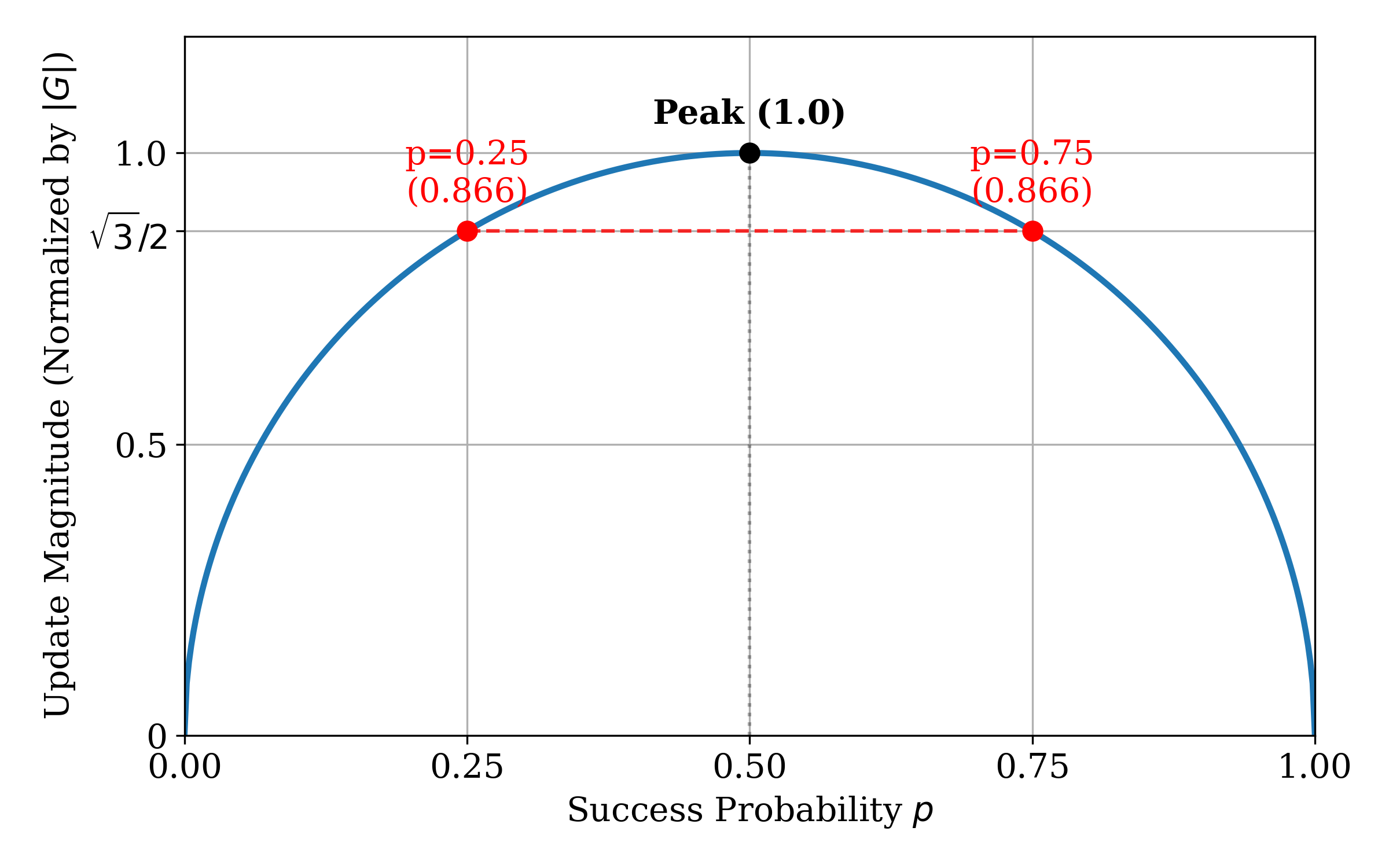}
\caption{\textbf{Visualization of Update Magnitude with Respect to Sample Difficulty.} The update magnitude reaches its peak at $p=0.5$, simple samples (e.g., $p=0.75$) and hard samples (e.g., $p=0.25$) exhibiting the same deviation from $p=0.5$ are assigned identical importance weights.}
\label{fig:sample-magnitude}
\end{figure}

\subsection{The Logits Update of Negative-Dominant Group in Behavior Space}
\label{sec:the logits of ndg}

\textbf{Theorem 1} (Behavior-Space Logit Update) 
\textit{Assume that the set of all possible behaviors is $B=\{b_{i}\}_{i=1}^{N}$, consisting of sampled set $G$ and unsampled set $U$. Given the intragroup advantage sum $C = \sum_{o_i \in G} A_{i}$, the behavior-level logit update of \(b_i\) under gradient ascent can be expressed as:}
\begin{equation}
\Delta h_{b_i}
=\eta\left[\mathbb I(b_i\in G)A_{b_i}-C\pi_{b_i}\right].
\end{equation}
\textit{where $\eta$ denotes the learning rate and $\pi_{b_i}$ is the model's current sampling probability in behavior space for path $b_i$.}

Down-weighting the advantages of correct paths leads to $C<0$, and we subsequently examine the dynamics of probability updates across the following cases.

Case A. For the sampled positive responses $b_{i} \in G_{pos}$:
\begin{equation}
\Delta h_{b_{i}} = \eta (A_{pos} - C \pi_{b_{i}}),
\end{equation}
where $A_{pos}>0$, $(- C \pi_{b_{i}})>0$ thus $\Delta h_{b_{i}}>0$. Hence, the sampling probability of a correct response will strictly increase.

Case B. For the sampled negative responses $b_{i} \in G_{neg}$:
\begin{equation}
\Delta h_{b_{i}} = \eta (A_{neg} - C \pi_{b_{i}}),
\end{equation}
where $A_{neg}<0$ but $(- C \pi_{b_{i}})>0$. Compared to GRPO($C=0$) which directly penalizes the incorrect responses, the logits of overconfident negative responses can even be increased. As reinforcement learning sharpens the output distribution~\cite{he2025rewarding,li2025jointly,lu2024distributionally,yue2025does}, overconfident erroneous responses occasionally displace correct ones in later stages, leading to a sharp surge in fully unsolved samples.


Case C. For the unsampled response $b_i \in U$:
\begin{equation}
\Delta h_{b_i} = \eta (- C \pi_{b_i})
\end{equation}
It can be observed that the output probabilities of all unsampled responses exhibit an increase, whereas they remain constant under GRPO. This phenomenon demonstrates that the Negative-Dominant group provides an exploration incentive, potentially facilitating the discovery of correct yet previously unsampled trajectories.

\section{Implementation Details}

\subsection{Dataset Introduction}

\paragraph{Mathematical Reasoning} 
We first conduct experiments on three widely recognized math benchmarks designed to evaluate the logical reasoning capabilities of LLM:
\begin{itemize}
    \item \textbf{MATH~\cite{hendrycks2021measuring}:} A comprehensive dataset containing challenging competition-level mathematics problems across various subjects.
    \item \textbf{AMC23:} Problems derived from the 2023 American Mathematics Competitions, representing high-school-level competitive reasoning.
    \item \textbf{AIME 2025:} The latest problems from the American Invitational Mathematics Examination, designed to test advanced problem-solving skills and long-chain reasoning.
\end{itemize}

\paragraph{Multi-modal Mathematical Reasoning} 
To validate the universality of our method in Vision-Language Models (VLMs), we extend our evaluation to three multi-modal mathematical reasoning datasets:
\begin{itemize}
    \item \textbf{Geo3K~\cite{lu2021inter}:} A dataset focused on geometry problem-solving that requires synchronized understanding of both visual diagrams and textual descriptions.
    \item \textbf{MathVerse~\cite{zhang2024mathverse}:} A comprehensive benchmark designed to evaluate visual mathematical reasoning across a wide array of subjects and visual patterns.
    \item \textbf{MathVision~\cite{wang2024measuring}:} A large-scale challenge consisting of diverse mathematical problems sourced from real-world competitions with complex visual contexts.
\end{itemize}

\paragraph{Medical Imaging Reasoning} 
Finally, to verify whether the proposed method enhances the model's ability to perceive and reason over fine-grained image details, we utilize the \textbf{HuatuoGPT-Vision}~\cite{chen2024huatuogpt} dataset. This benchmark is a curated and combined dataset from several publicly available medical VQA benchmarks, including VQA-RAD~\cite{lau2018dataset}, SLAKE~\cite{liu2021slake}, PathVQA~\cite{he2020pathvqa}, OmniMedVQA~\cite{hu2024omnimedvqa}, and PMC-VQA~\cite{zhang2023pmc}. It is specifically designed to assess medical domain expertise and visual grounding capabilities in complex clinical scenarios. Specifically, the training set comprises 600 MRI image-question pairs, while the evaluation is conducted on a diverse test suite consisting of 300 MRI, 300 CT, and 300 X-ray pairs following~\cite{pan2025medvlm}.

 \subsection{Hyperparameter Settings}
 \label{Hyperparameter Settings}

For all reinforcement learning experiments, responses were generated with a temperature of 1.0 and a maximum completion length of 2048 tokens. The learning rate is 1e-6. During evaluation, we used a generation temperature of 0.6, a top-p value of 0.95, and set the maximum new tokens to 4096. To ensure stable results, we perform 16
runs for AIME2025, and AMC23, and 4 runs for other benchmarks, reporting the average performance across the respective runs. Note that we did not use a dynamic sampling strategy in DAPO to ensure fair overhead. Our experiments are conducted over a single node with 8 NVIDIA H200 GPUs.

For experiments trained on the MATH dataset, we used the following system prompt to guide the model’s reasoning process: “Please reason step by step, and put your final answer within \textbackslash boxed\{\}.” The training batch size is set to 1024 with total 20 epochs. For each prompt, we generated 8 responses within the group. The reward was based on binary accuracy, where a correct final answer yielded a reward of 1 and an incorrect one yielded 0. 

For experiments trained on the Geo3K dataset, we used the following system prompt to guide the model’s reasoning process: “This is a multiple-choice question. You FIRST think about the reasoning process as an internal monologue and then provide the final choice from the given options. The reasoning process MUST BE enclosed within tags." The training batch size is set to 512 with total 20 epochs. The final answer MUST BE put in \textbackslash boxed\{\}.” For each prompt, we generated 5 responses within the group. The reward was based on 0.1*acc-reward + 0.9 * format-reward. As for the medical dataset, the training batch size is set to 32 with total 10 epochs, and the other settings are the same as those on Geo3K.


Our proposed A-GRAE method is highly accessible, requiring only a single hyperparameter: the scaling parameter $\alpha$ in~\cref{eq:attn}. In our experiments, $\alpha$ is set to $1$ for the Math dataset and $0.5$ for the Geo3K and Medical datasets. We also conduct the sensitivity analysis to $\alpha$ in~\cref{sec:sen-alpha}. Following prior works ~\cite{hochlehnert2025sober,chen2021evaluating,chen2025pass,zhu2025surprising}, we set our primary evaluation metric as Pass@$k$. Pass@$k$ estimates the probability of generating a correct solution within $k$ attempts. Unlike greedy decoding, Pass@$k$ provides a more reliable assessment of a model's potential and capability boundaries~\cite{hochlehnert2025sober,chen2021evaluating,chen2025pass,zhu2025surprising}. We employ the standard unbiased estimator, which generates $n$ responses ($n \ge k$) for each question $q$, counts the correct responses $c$, and computes the metric as follows:
\begin{equation}
    \text{Pass}@k = \mathbb{E}_{q \sim \mathcal{Q}} \left[ 1 - \frac{\binom{n-c}{k}}{\binom{n}{k}} \right]
\end{equation}

\section{Supplementary Experimental Results}

\subsection{Experimental Results of Control Experiment using Llama-3.2-3B-Instruct }
\label{Experimental Results using Llama-3.2-3B-Instruct}

It is observed that the empirical patterns on Llama-3.2-3B-Instruct remain largely consistent with those detailed in the main text. Specifically, down-weighting the positive advantages  at the group level significantly enhances reasoning performance, particularly in terms of the Pass@$k$ metric. Conversely, no universally optimal difficulty re-weighting strategy exists at the sample level. It is noteworthy, however, that when $k$ reaches 256, all evaluated methods underperform relative to the base model. This suggests that the backbone model itself often determines the efficacy of reinforcement learning, a finding consistent with recent literature~\cite{yeo2025demystifying,zhu2025surprising,wang2025octothinker}. Nevertheless, the Negative-Dominant group remains the most effective approach for preserving the fundamental capabilities of the base model.

    \begin{figure}[H] 
    \centering  
    \begin{subfigure}{0.32\textwidth}  
        \centering
        \includegraphics[width=\textwidth]{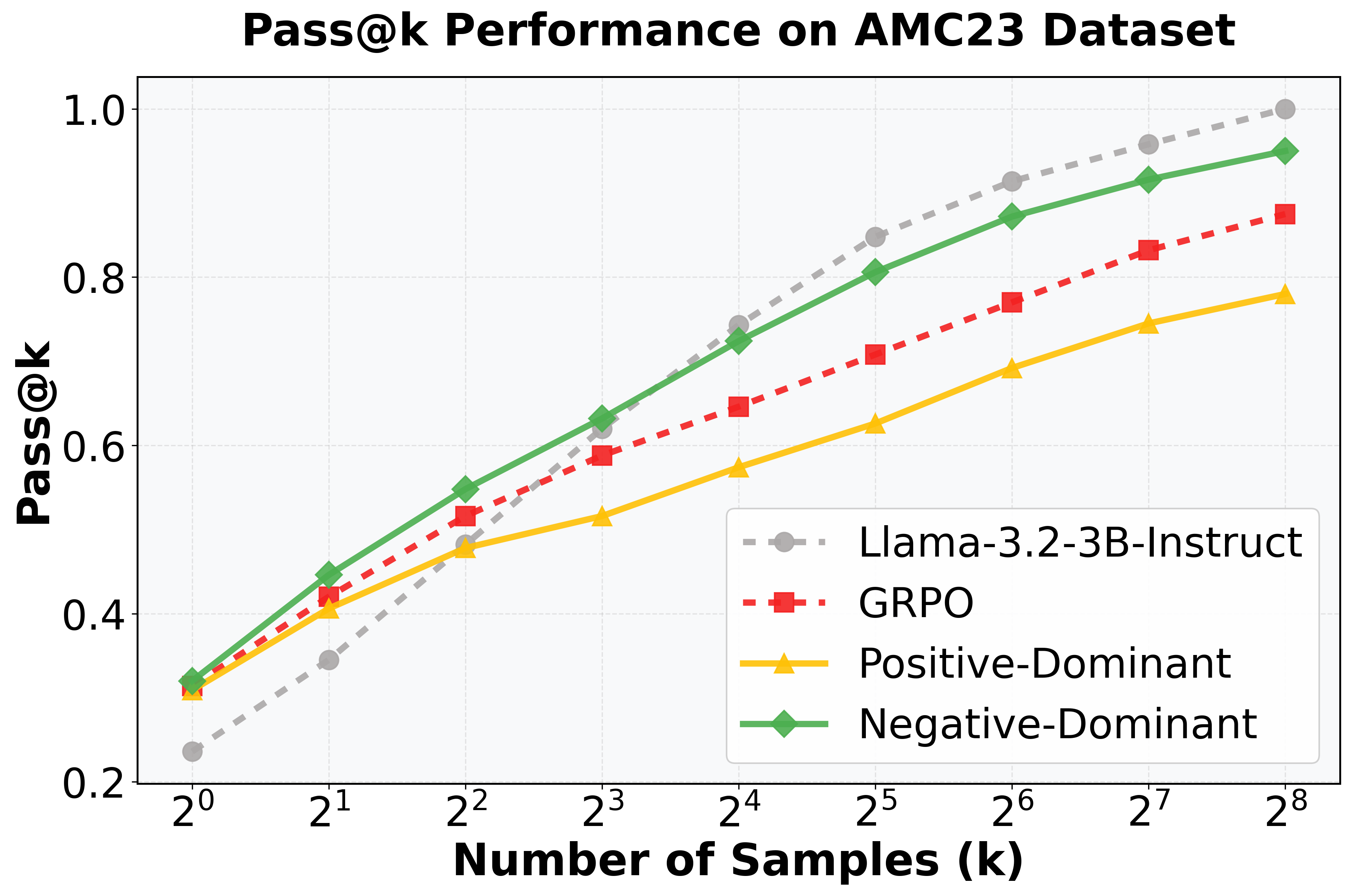}
        \label{subfig:llama-amc-g}
    \end{subfigure}
    \begin{subfigure}{0.32\textwidth} 
        \centering
        \includegraphics[width=\textwidth]{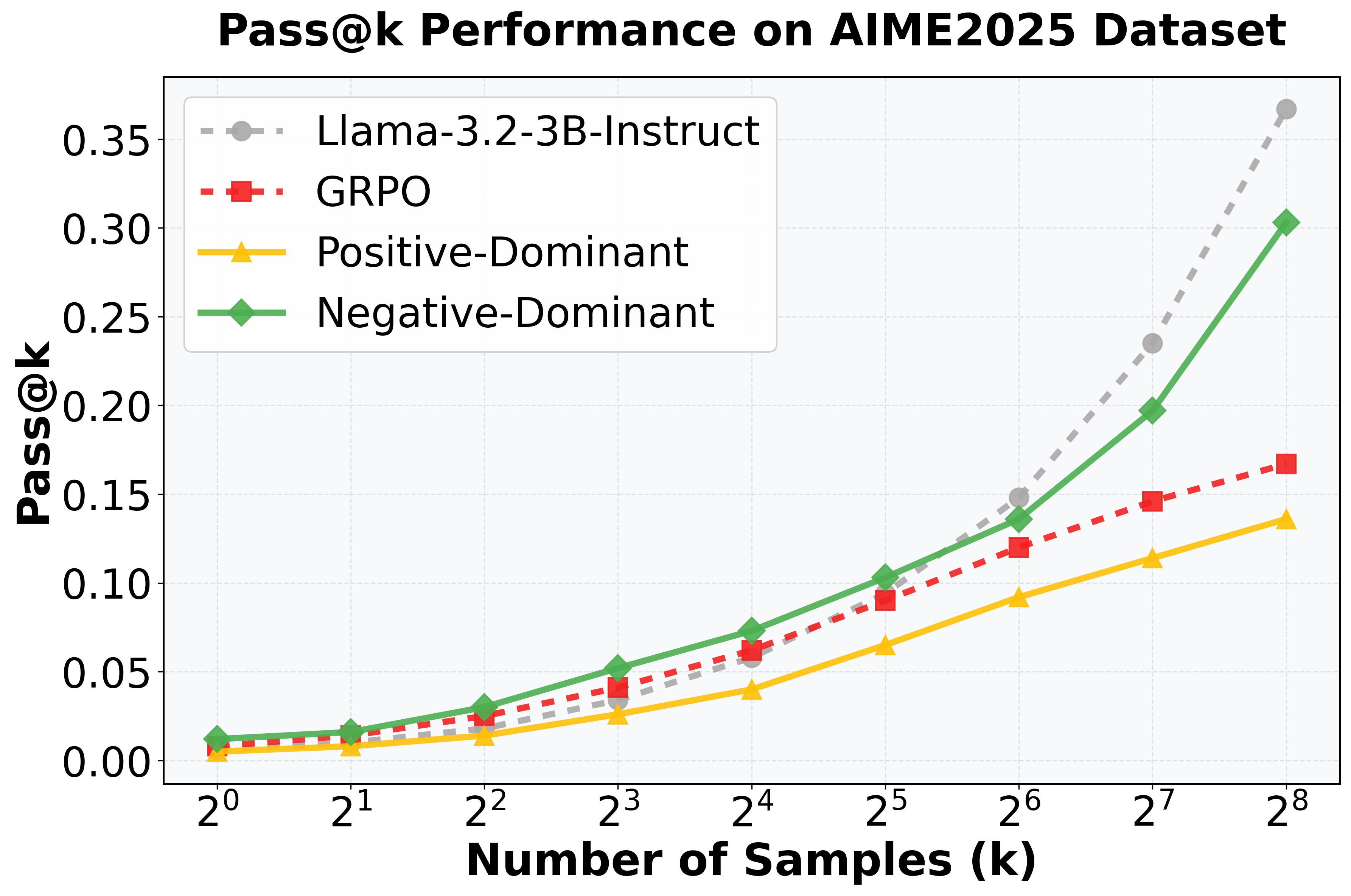}
        \label{subfig:llama-aime-g}
    \end{subfigure}
    \begin{subfigure}{0.32\textwidth} 
        \centering
        \includegraphics[width=\textwidth]{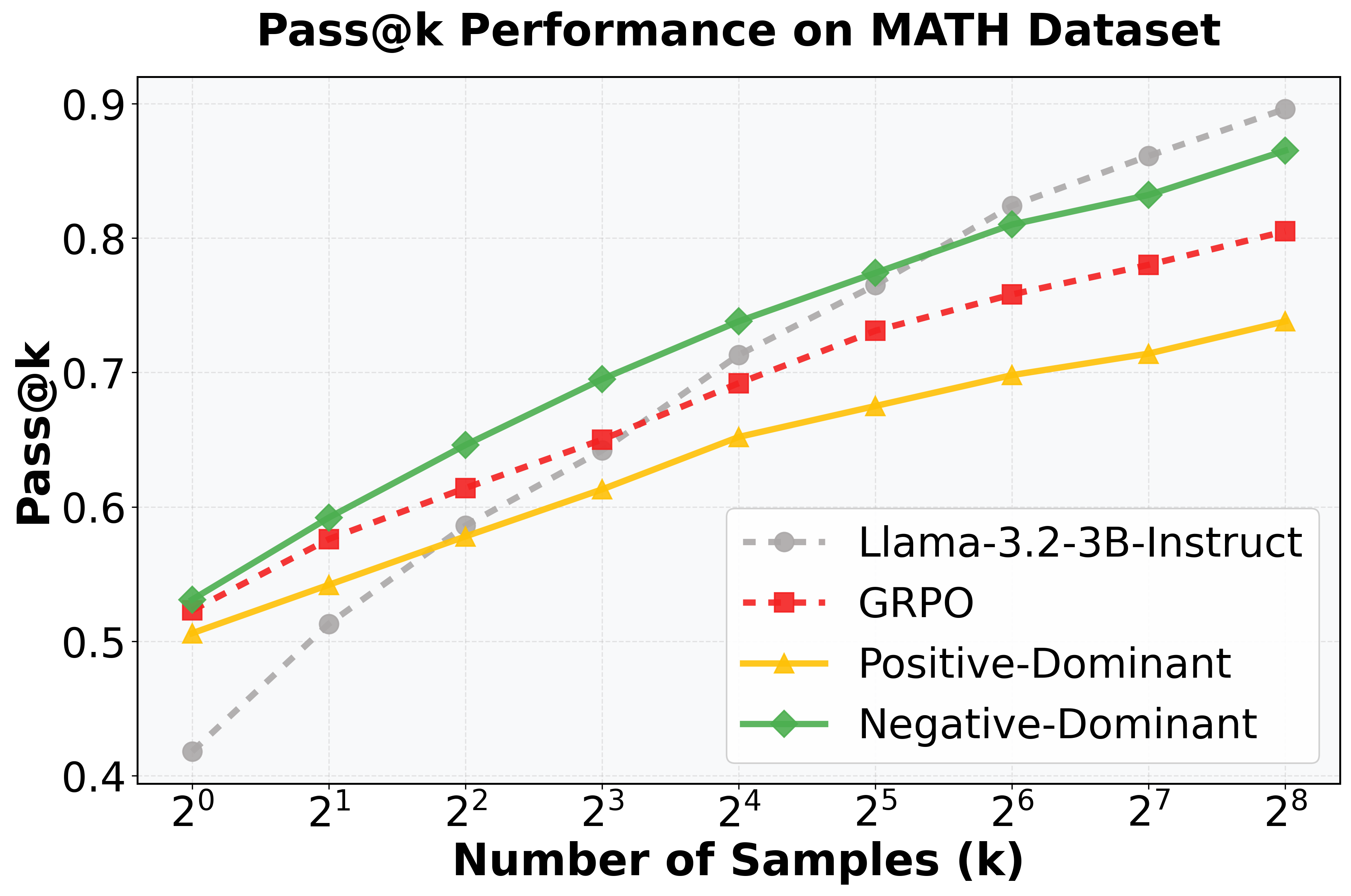}
    \end{subfigure}
    \vspace{-10pt}
    \caption{\textbf{Experimental results on breaking group-level symmetry using \texttt{Llama-3.2-3B-Instruct}.}  The performance is evaluated using Pass@$k$ ($k=\{1,2,4,8,16,32,64,128,256\}$) to provide a comprehensive analysis.}
    \label{fig:exp1-llama}
\end{figure}

    \begin{figure}[H] 
    \centering  
    \begin{subfigure}{0.32\textwidth}  
        \centering
        \includegraphics[width=\textwidth]{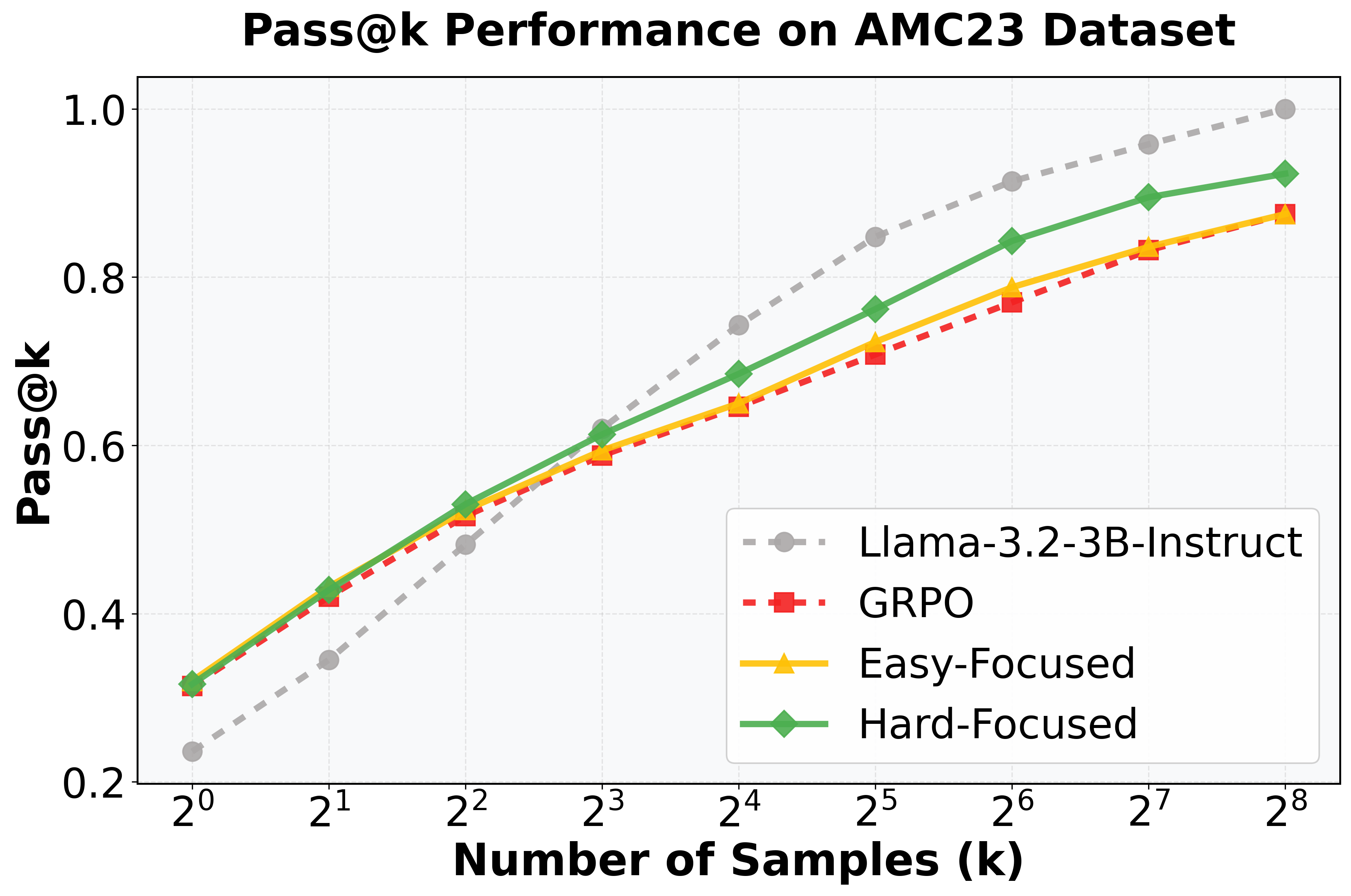}
        \label{subfig:llama-amc-s}
    \end{subfigure}
    \begin{subfigure}{0.32\textwidth} 
        \centering
        \includegraphics[width=\textwidth]{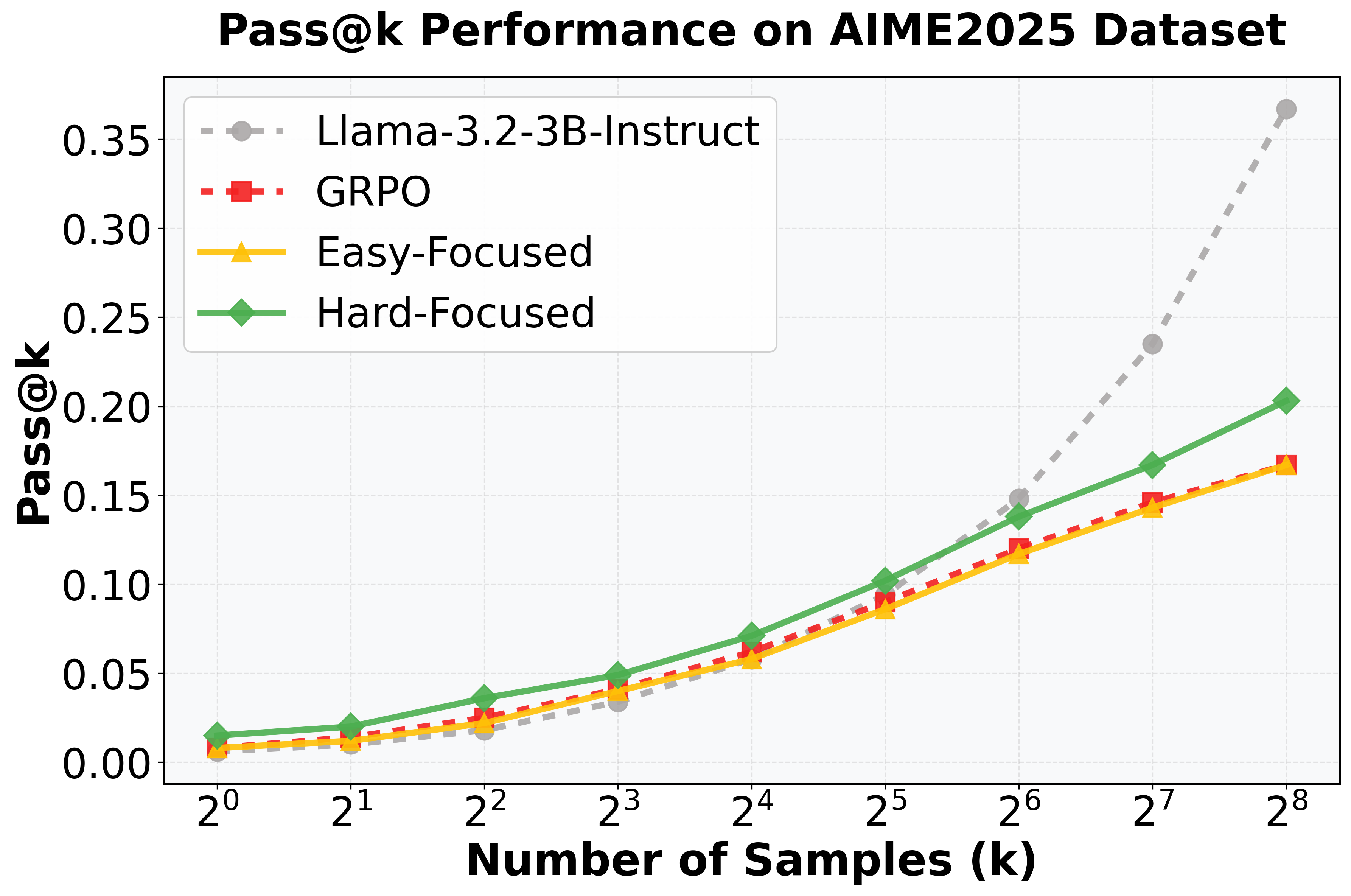}
        \label{subfig:llama-aime-s}
    \end{subfigure}
    \begin{subfigure}{0.32\textwidth} 
        \centering
        \includegraphics[width=\textwidth]{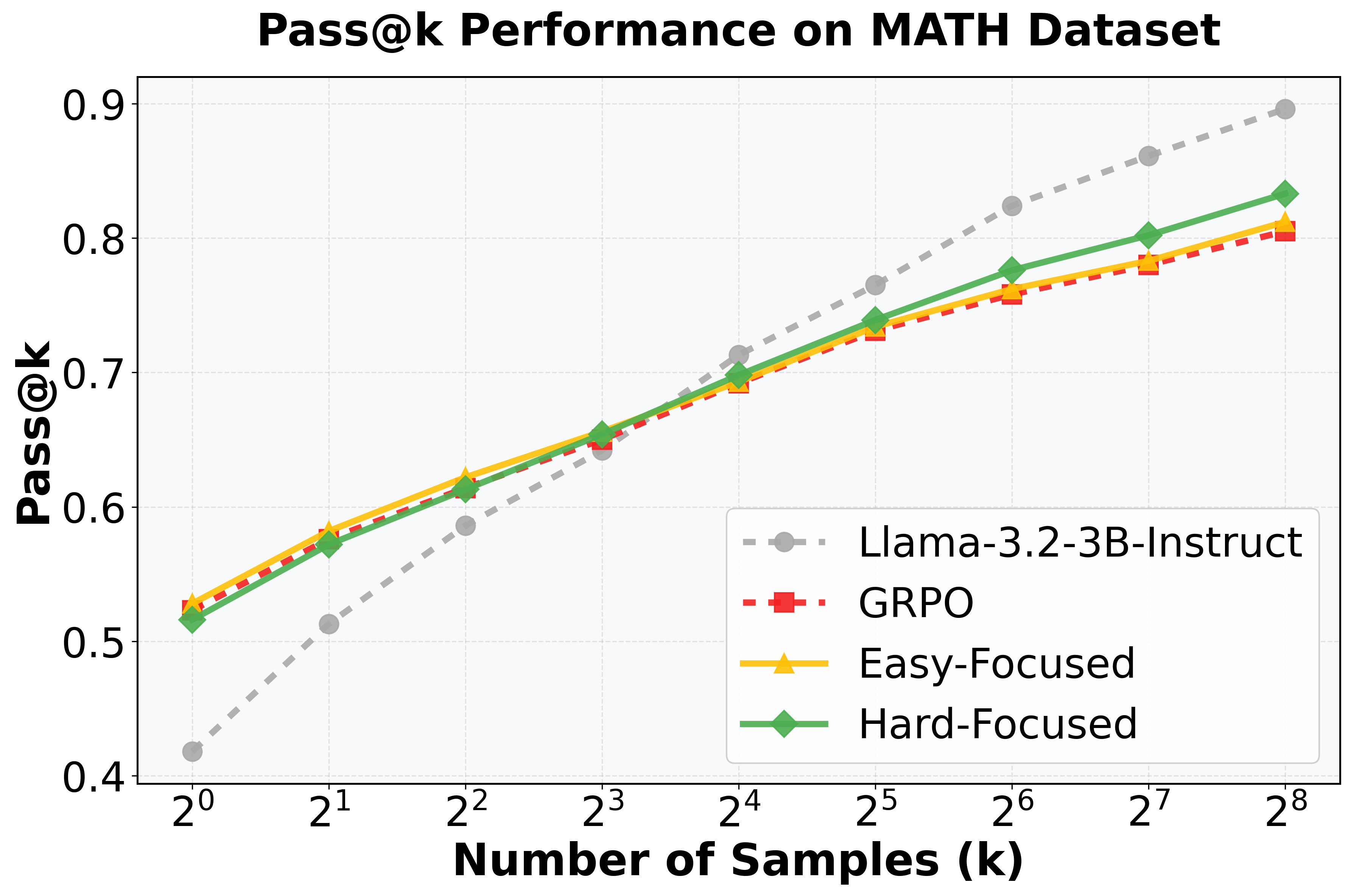}
    \end{subfigure}
    \vspace{-10pt}
    \caption{\textbf{Experimental results on breaking sample-level symmetry using \texttt{Llama-3.2-3B-Instruct}.} The performance is evaluated using Pass@$k$ ($k=\{1,2,4,8,16,32,64,128,256\}$) to provide a comprehensive analysis.}
    \label{fig:exp2-llama}
\end{figure}

\subsection{Supplementary Experimental Results using DeepSeek-R1-7B}
\label{sec:deepseek}

\begin{table*}[!h]
\centering
\caption{Pass@$k$ results on MATH, AIME 2025 and AMC23 with \texttt{DeepSeek-R1-7B}. \best{Bold} and \second{underlined} numbers denote the best and second-best results for each $k$. Specifically, We conduct 16 runs on AIME2025 and AMC23, and 4 runs on the remaining benchmarks.}
\label{tab:math-results-updated}
\small
\setlength{\tabcolsep}{3pt} 
\renewcommand{\arraystretch}{1.05}

\newcommand{\var}[1]{\hspace{1.5pt}{\scriptsize$\pm#1$}}

\resizebox{\textwidth}{!}{
\begin{tabular}{l cccccccccc}
\toprule
\best{Method} & \multicolumn{10}{c}{\best{Pass@$k$}} \\
\cmidrule(lr){2-11}
$k$ & 1 & 2 & 4 & 8 & 16 & 32 & 64 & 128 & 256 & \best{Avg.} \\
\midrule
\multicolumn{11}{c}{\best{MATH}} \\
Base Model & 84.3\var{0.5} & 89.6\var{0.6} & 92.8\var{0.5} & 94.8\var{0.4} & 96.0\var{0.4} & 96.8\var{0.3} & 97.4\var{0.3} & 97.8\var{0.3} & 98.0\var{0.2} & 94.2 \\
GRPO & 92.4\var{0.3} & 95.3\var{0.4} & 96.5\var{0.4} & 97.0\var{0.3} & 97.3\var{0.3} & 97.7\var{0.2} & 98.0\var{0.2} & 98.3\var{0.2} & 98.6\var{0.2} & 96.8 \\
\rowcolor{aliceblue} GRPO + A-GRAE & 92.9\var{0.2} & 95.7\var{0.3} & 96.9\var{0.3} & \second{97.5\var{0.3}} & 97.9\var{0.3} & 98.3\var{0.2} & \second{98.6\var{0.2}} & \best{99.0\var{0.2}} & \second{99.3\var{0.1}} & 97.3 \\
DAPO & 92.6\var{0.3} & 95.3\var{0.4} & 96.4\var{0.4} & 97.2\var{0.3} & 97.5\var{0.3} & 97.8\var{0.3} & 98.0\var{0.2} & 98.2\var{0.2} & 98.6\var{0.2} & 96.8 \\
\rowcolor{aliceblue} DAPO + A-GRAE & \second{93.0\var{0.2}} & \second{95.6\var{0.3}} & \second{97.0\var{0.3}} & \second{97.5\var{0.3}} & \second{98.1\var{0.2}} & \second{98.2\var{0.2}} & 98.4\var{0.2} & \second{98.7\var{0.2}} & 99.3\var{0.1} & 97.3 \\
Dr.GRPO & 92.8\var{0.3} & 95.6\var{0.4} & 96.8\var{0.4} & 97.3\var{0.3} & 97.7\var{0.3} & 98.0\var{0.2} & 98.3\var{0.2} & 98.5\var{0.2} & 98.8\var{0.2} & 97.2 \\
\rowcolor{aliceblue} Dr.GRPO + A-GRAE & \best{93.3\var{0.2}} & \best{96.0\var{0.3}} & \best{97.3\var{0.3}} & \best{97.8\var{0.2}} & \best{98.3\var{0.2}} & \best{98.6\var{0.2}} & \best{98.8\var{0.2}} & \best{99.0\var{0.1}} & \best{99.4\var{0.1}} & \best{97.6} \\
\midrule
\multicolumn{11}{c}{\best{AIME 2025}} \\
Base Model & 22.7\var{0.8} & 26.5\var{0.9} & 29.7\var{1.1} & 33.2\var{1.2} & 37.8\var{1.4} & 43.2\var{1.5} & 48.0\var{1.5} & 51.1\var{1.4} & 53.3\var{1.2} & 38.4 \\
GRPO & 28.5\var{0.5} & 35.0\var{0.7} & 41.4\var{0.9} & 47.7\var{1.1} & 53.1\var{1.2} & 57.6\var{1.4} & 62.1\var{1.3} & 67.2\var{1.3} & 73.3\var{1.1} & 51.8 \\
\rowcolor{aliceblue} GRPO + A-GRAE & 29.6\var{0.4} & 36.0\var{0.6} & 42.6\var{0.8} & 48.9\var{0.9} & 54.0\var{1.1} & 58.3\var{1.2} & 63.0\var{1.2} & 68.3\var{1.1} & 74.3\var{1.0} & 52.8 \\
DAPO & 29.2\var{0.6} & 35.5\var{0.8} & 42.2\var{1.0} & 48.5\var{1.2} & 53.3\var{1.3} & 57.7\var{1.5} & 62.5\var{1.4} & 67.6\var{1.4} & 73.6\var{1.2} & 52.2 \\
\rowcolor{aliceblue} DAPO + A-GRAE & \best{30.3\var{0.4}} & \best{36.8\var{0.6}} & \second{43.2\var{0.8}} & \second{49.3\var{0.9}} & \second{54.6\var{1.0}} & \second{58.6\var{1.2}} & \second{63.3\var{1.2}} & \second{68.4\var{1.1}} & \best{75.0\var{0.9}} & \best{53.3} \\
Dr.GRPO & 29.0\var{0.6} & 35.2\var{0.8} & 41.8\var{1.0} & 48.2\var{1.2} & 53.0\var{1.4} & 57.5\var{1.5} & 62.3\var{1.5} & 67.3\var{1.4} & 74.0\var{1.2} & 52.0 \\
\rowcolor{aliceblue} Dr.GRPO + A-GRAE & \second{29.8\var{0.4}} & \second{36.3\var{0.6}} & \best{43.3\var{0.8}} & \best{49.8\var{0.9}} & \best{55.2\var{1.1}} & \best{58.8\var{1.2}} & \best{63.5\var{1.2}} & \best{68.8\var{1.1}} & \second{74.6\var{1.0}} & \second{53.2} \\
\midrule
\multicolumn{11}{c}{\best{AMC23}} \\
Base Model & 64.2\var{0.6} & 73.0\var{0.7} & 80.1\var{0.8} & 85.5\var{0.9} & 89.8\var{0.9} & 92.8\var{0.8} & 95.2\var{0.7} & 97.4\var{0.5} & \best{100.0\var{0.0}} & 86.4 \\
GRPO & 82.8\var{0.4} & 89.3\var{0.5} & 92.3\var{0.6} & 94.0\var{0.7} & 95.2\var{0.7} & 96.1\var{0.6} & 97.2\var{0.5} & 98.8\var{0.4} & \best{100.0\var{0.0}} & 94.0 \\
\rowcolor{aliceblue} GRPO + A-GRAE & 84.2\var{0.3} & 90.6\var{0.4} & 93.5\var{0.5} & 95.4\var{0.6} & 96.6\var{0.6} & 97.7\var{0.6} & 98.0\var{0.5} & 99.3\var{0.3} & \best{100.0\var{0.0}} & 95.0 \\
DAPO & 83.3\var{0.4} & 89.8\var{0.5} & 92.6\var{0.6} & 94.3\var{0.7} & 95.8\var{0.7} & 96.5\var{0.7} & 97.6\var{0.6} & 99.2\var{0.4} & \best{100.0\var{0.0}} & 94.3 \\
\rowcolor{aliceblue} DAPO + A-GRAE & \best{84.6\var{0.3}} & \best{91.6\var{0.4}} & \best{94.2\var{0.5}} & \best{96.3\var{0.6}} & \second{97.0\var{0.6}} & \second{97.8\var{0.6}} & \second{98.2\var{0.5}} & \best{99.5\var{0.3}} & \best{100.0\var{0.0}} & \best{95.5} \\
Dr.GRPO & 83.0\var{0.5} & 89.6\var{0.6} & 92.3\var{0.7} & 93.8\var{0.8} & 95.6\var{0.8} & 96.3\var{0.7} & 97.2\var{0.6} & 98.2\var{0.5} & \best{100.0\var{0.0}} & 94.0 \\
\rowcolor{aliceblue} Dr.GRPO + A-GRAE & \second{84.5\var{0.3}} & \second{91.3\var{0.4}} & \second{93.6\var{0.5}} & \second{96.0\var{0.6}} & \best{97.2\var{0.6}} & \best{98.0\var{0.6}} & \best{98.5\var{0.5}} & \second{99.3\var{0.4}} & \best{100.0\var{0.0}} & \second{95.4} \\
\bottomrule
\end{tabular}}
\end{table*}

\subsection{Training Collapse of Negative-Dominant Group}
\label{Training Collapse of Negative-Dominant Group}

\begin{figure}[!htbp] 
    \centering  
    \begin{subfigure}{0.45\textwidth}  
        \centering
        \includegraphics[width=\textwidth]{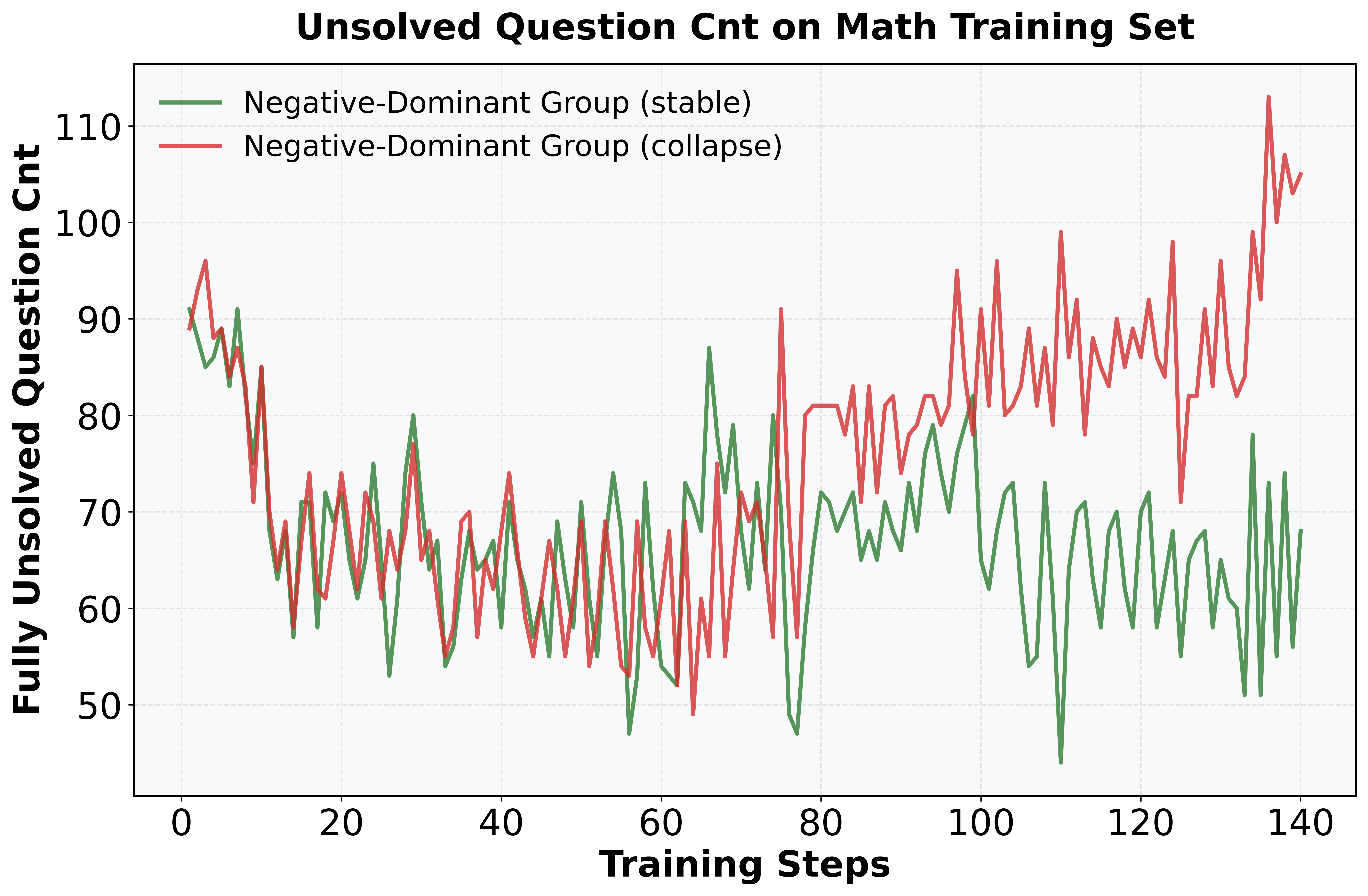}
        \label{subfig:unsolved}
    \end{subfigure}
    \begin{subfigure}{0.45\textwidth} 
        \centering
        \includegraphics[width=\textwidth]{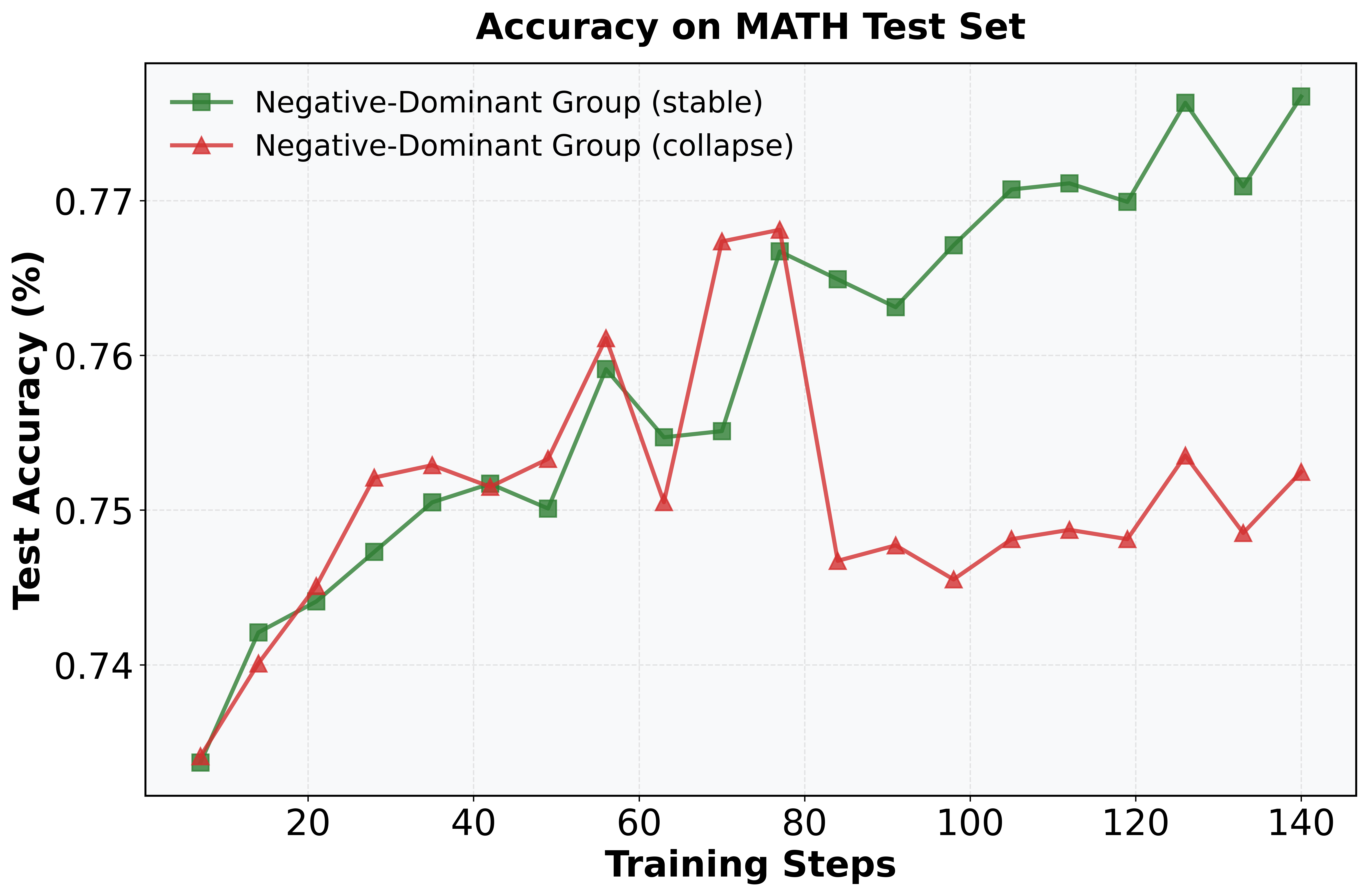}
        \label{subfig:testacc}
    \end{subfigure}
    \vspace{-15pt}
    \caption{\textbf{Training collapse of Negative-Dominant group during training with \texttt{Qwen2.5-Math-7B}.} \textbf{Left Part:} The unsolved question count suddenly increase in the later stage. \textbf{Right Part:} The collapse leads to the significant performance decrease.}
    \label{fig:unstable}
\end{figure}

To further investigate the reliability of the Negative-Dominant configuration, we conducted multiple training runs under identical hyperparameter settings. As illustrated in~\cref{fig:unstable}, we observe a striking stochastic instability within this specific group. While one trial remains relatively stable throughout the training process, the other undergoes a catastrophic training collapse after approximately step 78. This divergence highlights that the Negative-Dominant approach is highly sensitive to initial conditions or sampling noise. In the "collapse" instance, the number of fully unsolved questions exhibits high-amplitude oscillations and a persistent upward trend, eventually peaking at over 110. This suggests that without proper regularization, the Negative-Dominant objective can easily lead the model into a divergent optimization path, where erroneous trajectories reinforce themselves and eventually overwhelm the model's reasoning capabilities. Such unpredictable behavior underscores the necessity of balancing negative trajectory weights to ensure consistent convergence.

\subsection{Experimental Results in Further Analysis}

\begin{table*}[!h]
\centering
\caption{Ablation studies on MATH, AIME 2025 and AMC23 with \texttt{Qwen2.5-Math-7B}.}
\label{tab:ablation}
\small 
\setlength{\tabcolsep}{8pt} 
\renewcommand{\arraystretch}{1.1} 
\resizebox{\textwidth}{!}{
\begin{tabular}{l ccccccccc}
\toprule
\best{Method} & \multicolumn{9}{c}{\best{Pass@$k$}} \\
\cmidrule(lr){2-10}
$k$ & 1 & 2 & 4 & 8 & 16 & 32 & 64 & 128 & 256 \\
\midrule
\multicolumn{10}{c}{\best{MATH}} \\
Base Model & 63.4 & 74.8 & 83.2 & 88.6 & 91.2 & 93.4 & 94.1 & 95.0 & 96.3 \\
GRPO       & 76.5 & 82.3 & 86.1 & 88.8 & 90.3 & 92.6 & 93.5 & 93.9 & 95.0 \\
\rowcolor{aliceblue} GRPO + A-GRAE (sample level)  & \second{77.8} & \second{84.2} & \second{88.4} & 90.5 & 92.1 & 94.0 & 94.3 & 94.8 & 96.0 \\
\rowcolor{aliceblue} GRPO + A-GRAE (group level)  & 77.6 & 84.0 & 88.1 & \best{91.2} & \best{93.0} & \second{94.3} & \best{95.3} & \best{95.6} & \best{97.0} \\
\rowcolor{aliceblue} GRPO + A-GRAE (full)  & \best{78.3} & \best{85.0} & \best{89.2} & \second{91.0} & \second{92.5} & \best{94.6} & \second{95.0} & \second{95.5} & \second{96.5} \\
\midrule
\multicolumn{10}{c}{\best{AIME 2025}} \\
Base Model & 6.1 & 9.9 & 14.4 & 19.3 & 24.4 & 29.1 & 33.4 & 39.2 & 46.7 \\
GRPO       & 10.3 & 14.3 & 18.7 & 23.1 & 27.5 & 31.8 & 36.1 & 40.8 & 46.7 \\
\rowcolor{aliceblue} GRPO + A-GRAE (sample level)  & 10.9 & 15.1 & 19.6 & 24.1 & 28.0 & 33.5 & 40.0 & 47.0 & 52.3 \\
\rowcolor{aliceblue} GRPO + A-GRAE (group level)  & \second{11.0} & \second{15.4} & \best{20.3} & \second{24.5} & \second{28.3} & \second{33.9} & \best{41.0} & \best{49.6} & \best{60.0} \\
\rowcolor{aliceblue} GRPO + A-GRAE (full)  & \best{11.3} & \best{15.6} & \second{19.8} & \best{24.7} & \best{28.6} & \best{34.1} & \second{39.2} & \second{47.8} & \second{56.7} \\
\midrule
\multicolumn{10}{c}{\best{AMC23}} \\
Base Model & 40.6 & 55.3 & 68.6 & 78.6 & 85.0 & 89.4 & 93.4 & \best{97.3} & \best{100.0} \\
GRPO       & 59.2 & 66.7 & 72.1 & 76.4 & 80.6 & 84.8 & 88.3 & 90.8 & 92.5 \\
\rowcolor{aliceblue} GRPO + A-GRAE (sample level)  & \second{62.3} & \best{70.4} & 76.8 & \second{83.0} & 86.3 & 90.4 & 93.2 & 94.3 & 95.0 \\
\rowcolor{aliceblue} GRPO + A-GRAE (group level)  & 61.4 & 69.8 & \second{77.0} & 82.3 & \second{87.0} & \second{90.8} & \second{94.4} & \best{97.3} & \best{100.0} \\
\rowcolor{aliceblue} GRPO + A-GRAE (full)  & \best{62.6} & \second{70.0} & \best{77.5} & \best{83.7} & \best{88.2} & \best{92.0} & \best{95.1} & \second{96.8} & \second{97.5} \\
\bottomrule
\end{tabular}}
\end{table*}

\begin{figure}[!h] 
    \centering  
    \begin{subfigure}{0.32\textwidth}  
        \centering
        \includegraphics[width=\textwidth]{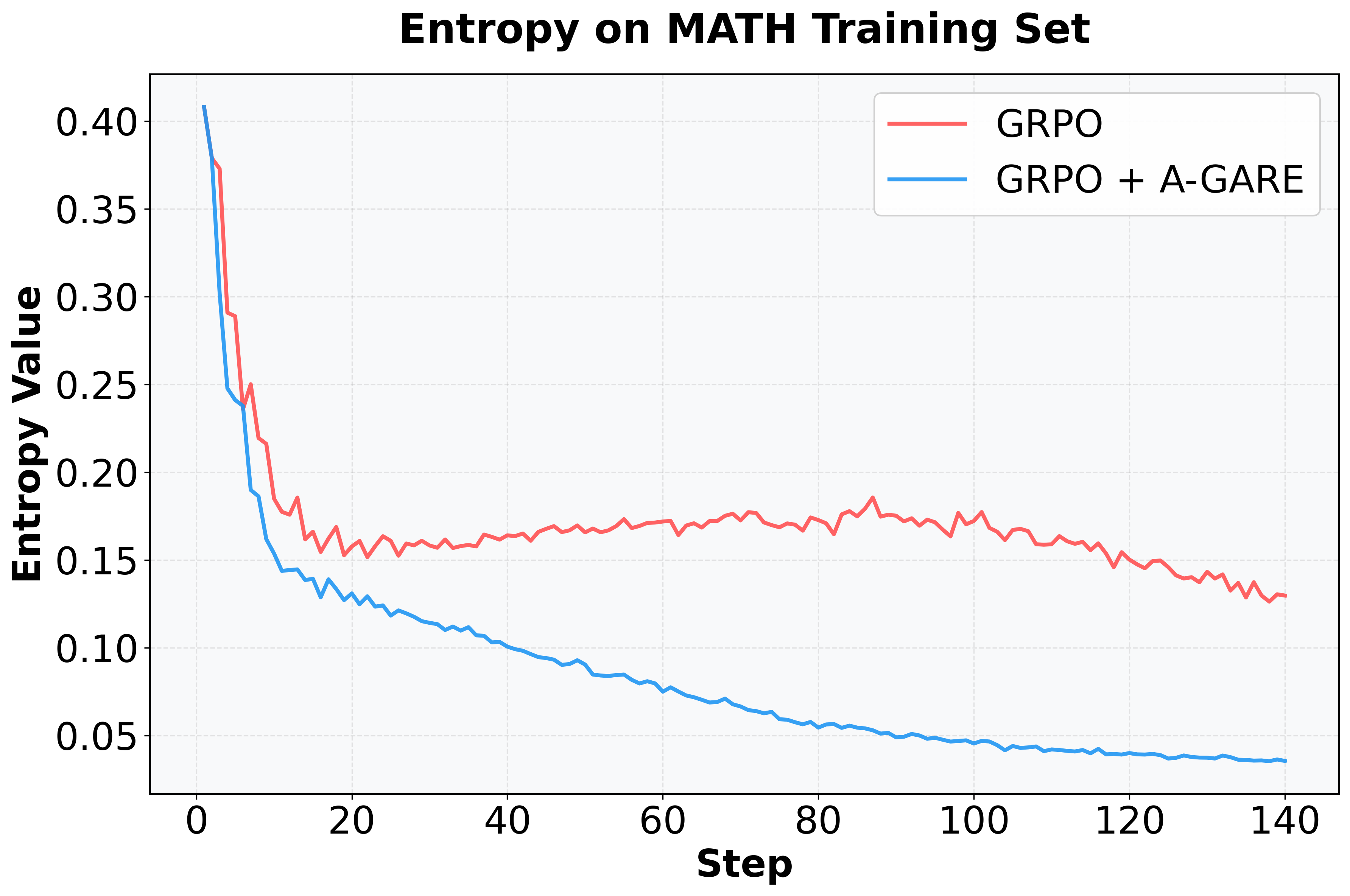}
        \label{subfig:ent-train}
    \end{subfigure}
    \begin{subfigure}{0.32\textwidth} 
        \centering
        \includegraphics[width=\textwidth]{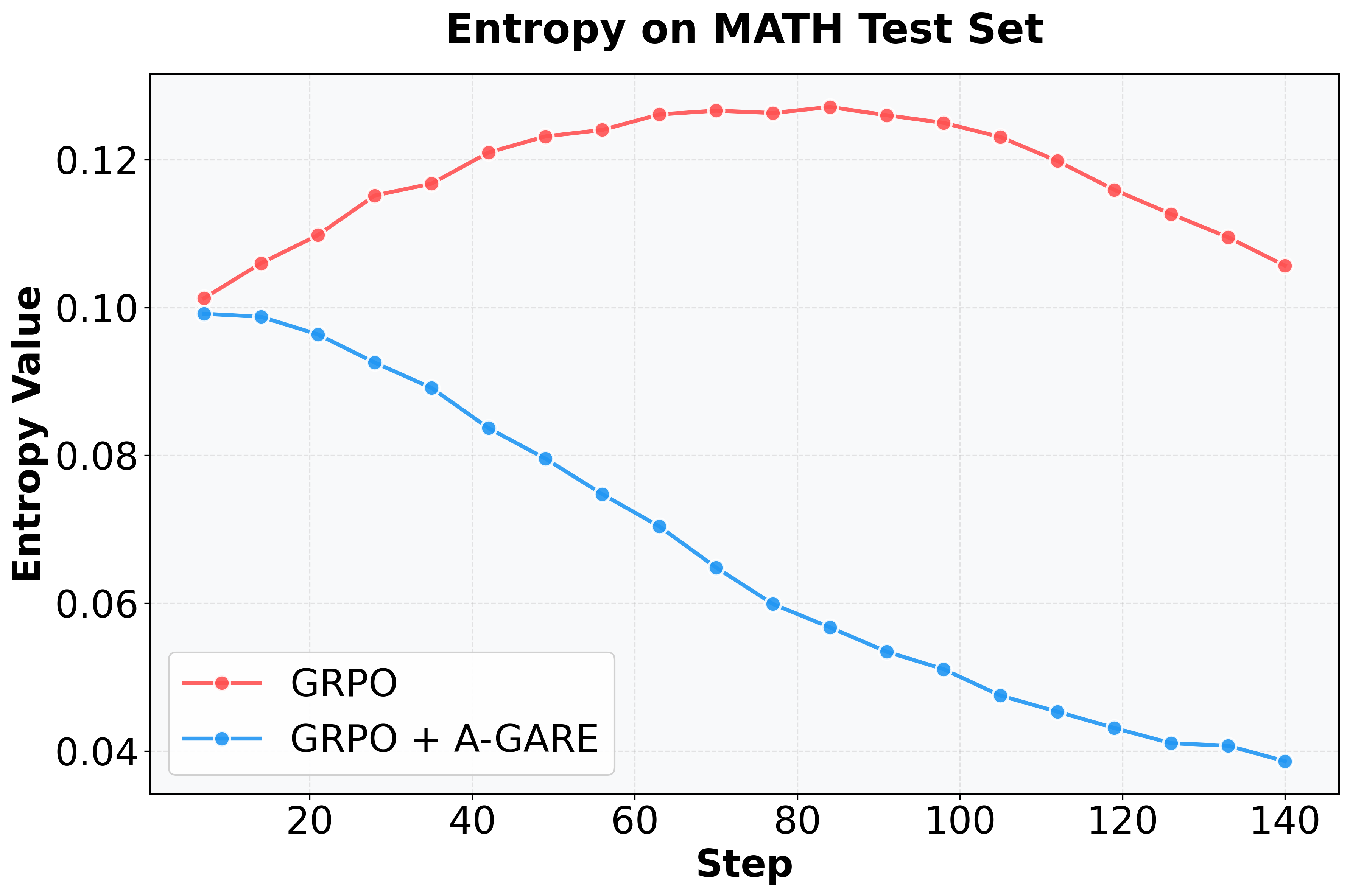}
        \label{subfig:ent-test}
    \end{subfigure}
    \begin{subfigure}{0.32\textwidth} 
        \centering
        \includegraphics[width=\textwidth]{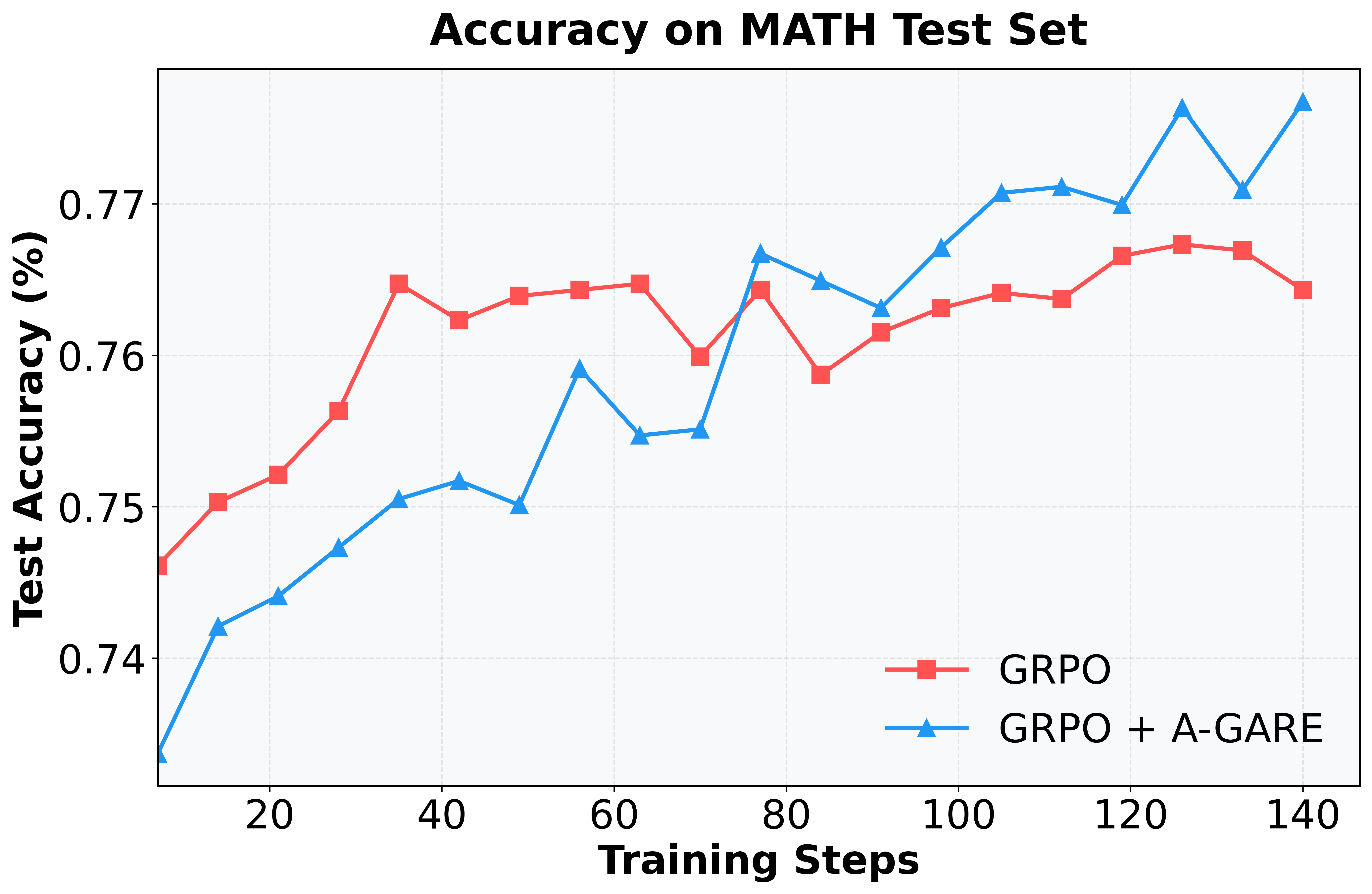}
        \label{subfig:dyn-acc}
    \end{subfigure}
    \vspace{-10pt}
    \caption{\textbf{Training dynamics using \texttt{Qwen2.5-Math-7B}.} \textbf{Left Part:} Model's entropy on the Math training set. \textbf{Center Part:}  Model's entropy (actor entropy loss) on the Math test set. \textbf{Right Part:} The greedy decoding accuracy on the Math test set.}
    \label{fig:dynamics}
\end{figure}

\subsection{Comparison Experiments with Entropy Promoting Techniques}
\label{sec:Comparison Experiments with Entropy Promoting Techniques}

To verify whether the gains of our approach stem merely from a generic entropy-promoting effect, we conduct an additional entropy-matched control experiment. We implement an "Entropy-Promoting" baseline by tuning the \texttt{entropy\_coeff} in the official \textit{verl} framework. Specifically, by setting the coefficient to $8\mathrm{e}{-3}$, we achieve a training-set entropy of $0.758$ at step 105, which closely matches that of the Negative-Dominant group ($0.742$). As illustrated in Table~\ref{tab:entropy_comparison}, simply increasing the entropy does not improve reasoning performance. Despite explicitly encouraging higher entropy, the entropy-promoting baseline consistently underperforms vanilla GRPO across all $k$ values. Upon further inspection, we observe that indiscriminate entropy promotion often leads to the "repetitive-output" phenomenon, where higher entropy corresponds to low-quality or redundant trajectories rather than meaningful diversity~\cite{jiang2025rethinking}. In contrast, our Negative-Dominant strategy, which breaks the symmetry in GRAE, consistently achieves the highest Pass@$k$ scores, reaching $92.8\%$ at $k=16$ and maintaining superior scaling up to $k=256$. This suggests that effective exploration must be encouraged fundamentally at the trajectory level through symmetry breaking, rather than through global entropy regularization.

\begin{table}[h]
\centering
\caption{Comparison of MATH Pass@$k$ (\%) between Negative-Dominant group and entropy-promoting techniques.}
\label{tab:entropy_comparison}
\begin{tabular}{lcccccc}
\toprule
Method & $k=1$ & $k=2$ & $k=4$ & $k=8$ & $k=16$ & $k=256$ \\ \midrule
GRPO (Baseline) & 76.5 & 82.3 & 86.1 & 88.8 & 90.3 & 92.4 \\
Negative-Dominant & \textbf{77.2} & \textbf{83.8} & \textbf{87.6} & \textbf{90.4} & \textbf{92.8} & \textbf{94.5} \\
GRPO + Entropy-Promoting & 75.8 & 80.3 & 83.6 & 86.5 & 88.4 & 89.0\\ \bottomrule
\end{tabular}
\end{table}

\subsection{Comprehensive Statistical Significance Analysis over Baselines}
\label{sec:detailed_p_value}

To assess A-GRAE's robustness across computational budgets, we conduct a fine-grained significance analysis using Welch's t-test across three primary baselines (GRPO, DAPO, and Dr.GRPO) at each interval $k \in [1, 256]$. As detailed in Table~\ref{tab:detailed_p_values}, A-GRAE consistently yields $p < 0.05$ across all benchmarks, confirming that its gains are statistically significant rather than stochastic fluctuations. Notably, on the MATH benchmark, Dr.GRPO + A-GRAE maintains $p < 0.005$ throughout scaling and retains a significant lead at $k=256$ ($p=0.0038$), demonstrating its efficacy in mitigating the capability boundary shrinkage observed in vanilla GRPO. For the high-variance AIME 2025 and AMC23 tasks, A-GRAE provides even more pronounced benefits; specifically, DAPO + A-GRAE achieves $p < 0.001$ across nearly all $k$ intervals on AIME 2025. These consistent results across diverse baselines and datasets underscore the universal efficacy of breaking advantage symmetry to facilitate stable and efficient policy optimization during multi-attempt scaling.

\begin{table*}[h]
\centering
\caption{Detailed $p$-values across scaling intervals ($k=1$ to $k=256$). $p$-values are calculated comparing the A-GRAE enhanced version against its respective vanilla baseline.}
\label{tab:detailed_p_values}
\small
\resizebox{\textwidth}{!}{
\begin{tabular}{lccccccccc}
\toprule
\textbf{Comparison Pair} & $k=1$ & $k=2$ & $k=4$ & $k=8$ & $k=16$ & $k=32$ & $k=64$ & $k=128$ & $k=256$ \\ \midrule
\rowcolor{gray!10} \multicolumn{10}{l}{\textbf{MATH Benchmark}} \\
GRPO + A-GRAE vs. GRPO & .0032 & .0018 & .0025 & .0041 & .0084 & .0115 & .0128 & .0155 & .0121 \\
DAPO + A-GRAE vs. DAPO & .0015 & .0022 & .0048 & .0055 & .0072 & .0098 & .0105 & .0112 & .0108 \\
Dr.GRPO + A-GRAE vs. Dr.GRPO & .0024 & .0009 & .0011 & .0022 & .0015 & .0028 & .0035 & .0042 & .0038 \\ \midrule
\rowcolor{gray!10} \multicolumn{10}{l}{\textbf{AIME 2025 Benchmark}} \\
GRPO + A-GRAE vs. GRPO & .0125 & .0098 & .0112 & .0105 & .0142 & .0128 & .0115 & .0092 & .0085 \\
DAPO + A-GRAE vs. DAPO & .0016 & .0012 & .0008 & .0005 & .0008 & .0006 & .0004 & .0005 & .0003 \\
Dr.GRPO + A-GRAE vs. Dr.GRPO & .0055 & .0042 & .0068 & .0085 & .0102 & .0092 & .0115 & .0088 & .0076 \\ \midrule
\rowcolor{gray!10} \multicolumn{10}{l}{\textbf{AMC23 Benchmark}} \\
GRPO + A-GRAE vs. GRPO & .0045 & .0038 & .0042 & .0051 & .0048 & .0062 & .0075 & .0082 & .0065 \\
DAPO + A-GRAE vs. DAPO & .0012 & .0008 & .0005 & .0006 & .0004 & .0005 & .0011 & .0015 & .0008 \\
Dr.GRPO + A-GRAE vs. Dr.GRPO & .0008 & .0012 & .0009 & .0007 & .0011 & .0015 & .0008 & .0022 & .0015 \\ \bottomrule
\end{tabular}}
\vspace{2pt}
\\ \small \textit{Note: All reported $p$-values $< 0.05$ are considered statistically significant.}
\end{table*}

\section{Sensitivity to Hyperparameter \texorpdfstring{$\alpha$}{alpha}}
\label{sec:sen-alpha}

We evaluate the sensitivity of hyperparameter $\alpha$ using the Qwen2.5-Math-7B backbone under the GRPO training paradigm. As shown in Table~\ref{tab:alpha_sensitivity}, the results indicate that our method is fairly robust to $\alpha$. The influence of $\alpha$ can be theoretically understood through the reformulated advantage estimation: $A_i \cdot \min(1, w_s/\alpha)$.  Here, $\alpha$ controls the degree of suppression on positive advantages: when $\alpha$ is small, the latter term always take 1, so the method behaves similarly to GRPO; as $\alpha$ increases, the positive advantage is gradually attenuated, and in the limit where $\alpha \to \infty$, it is nearly removed. Although excessively large $\alpha$ causes some performance degradation, learning remains effective overall, consistent with previous observations (GRPO-N~\cite{yao2025debate}), which removes the positive advantage in GRPO entirely.

\begin{table}[h]
\centering
\caption{Sensitivity analysis of $\alpha$ on Qwen2.5-Math-7B. Performance is measured by Pass@1 (\%).}
\label{tab:alpha_sensitivity}
\begin{tabular}{lccc}
\toprule
Method / Setting & MATH & AMC23 & AIME2025 \\ \midrule
GRPO (Baseline)  & 76.5 & 59.2  & 10.3     \\ \midrule
$\alpha = 0.25$  & 77.8 & 61.8  & 10.9     \\
$\alpha = 0.5$   & 77.8 & 61.8  & 11.0     \\
$\alpha = 1.0$   & \textbf{78.3} & 62.6  & \textbf{11.3}     \\
$\alpha = 2.0$   & 78.0 & \textbf{62.8}  & \textbf{11.3}     \\
$\alpha = 5.0$   & 76.8 & 62.3  & 10.8     \\ \bottomrule
\end{tabular}
\end{table}

\section{Limitations}

Despite the theoretical insights and performance gains provided by A-GRAE, several limitations remain: While A-GRAE effectively mitigates the shrinkage of capability boundaries, it remains bounded by the inherent reasoning ceilings of the base model backbones, meaning it cannot fully compensate for fundamental knowledge gaps in smaller models. This study also primarily focuses on Reinforcement Learning with Verifiable Rewards (RLVR) using binary reward signals; therefore, the extent to which implicit advantage symmetry affects scenarios with dense, partial, or human-preference-based rewards remains to be explored in future work. Finally, although the method retains the efficiency of a critic-free architecture, finding the optimal balance between exploration and convergence in highly complex domains may still require extended training schedules.



\end{document}